\documentclass[conference]{IEEEtran}
\IEEEoverridecommandlockouts
\usepackage{cite}
\usepackage{amsmath,amssymb,amsfonts}
\usepackage{algorithmic}
\usepackage{algorithm}
\usepackage{graphicx}
\usepackage{textcomp}
\usepackage{xcolor}
\usepackage{subcaption}
\usepackage{url}
\usepackage{booktabs}
\usepackage{xcolor}
\usepackage{comment}
\usepackage[normalem]{ulem}
\newcommand{\InlineComment}[1]{\STATE \textcolor{gray}{\# \textit{#1}}}

\newcommand{\bigO}{\mathcal{O}}
\def\BibTeX{{\rm B\kern-.05em{\sc i\kern-.025em b}\kern-.08em
    T\kern-.1667em\lower.7ex\hbox{E}\kern-.125emX}}
\begin{document}

\title{Enhanced Soups for Graph Neural Networks\\
}

\author{
    \IEEEauthorblockN{Joseph Zuber, Aishwarya Sarkar, Joseph Jennings, Ali Jannesari}
    \IEEEauthorblockA{Iowa State University, Ames, IA, USA}
    \{zubes, asarkar1, jtj9, jannesar\}@iastate.edu
}

% \author{
% \IEEEauthorblockN{Joseph Zuber}
% \IEEEauthorblockA{
% \textit{Iowa State University}\\
% Ames, USA \\
% zubes@iastate.edu}
% \and
% \IEEEauthorblockN{Aishwarya Sarkar}
% \IEEEauthorblockA{
% \textit{Iowa State University}\\
% Ames, USA \\
% asarkar1@iastate.edu}
% \and
% \IEEEauthorblockN{Joseph Jennings}
% \IEEEauthorblockA{
% \textit{Iowa State University}\\
% Ames, USA \\
% jtj9@iastate.edu}
% \and
% \IEEEauthorblockN{Ali Jannesari}
% \IEEEauthorblockA{
% \textit{Iowa State University}\\
% Ames, USA \\
% jannesar@iastate.edu}
% }

\maketitle

\begin{abstract}
Graph Neural Networks (GNN) have demonstrated state-of-the-art performance in numerous scientific and high-performance computing (HPC) applications. Recent work suggests that ``souping" (combining) individually trained GNNs into a single model can improve performance without increasing compute and memory costs during inference. However, existing souping algorithms are often slow and memory-intensive, which limits their scalability.

We introduce \textit{Learned Souping} for GNNs, a gradient-descent-based souping strategy that substantially reduces time and memory overhead compared to existing methods. Our approach is evaluated across multiple Open Graph Benchmark (OGB) datasets and GNN architectures, achieving up to 1.2\% accuracy improvement and 2.1X speedup. Additionally, we propose \textit{Partition Learned Souping}, a novel partition-based variant of learned souping that significantly reduces memory usage. On the ogbn-products dataset with GraphSAGE, partition learned souping achieves a 24.5X speedup and a 76\% memory reduction without compromising accuracy.
% , and test it against several large datasets using state-of-the-art model architectures. This approach sees an impressive 1.2\% accuracy gain and 2.1X speedup on the Reddit dataset with the GAT architecture. We also introduce a novel partition-learned variant of this approach, which can decrease GPU memory usage significantly. On the ogbn-products dataset with the GraphSAGE architecture, this partition-learned variant achieves a 24.5X speedup and a 76\% memory reduction without degrading the accuracy.
\end{abstract}


\section{Introduction}
\begin{comment}
Graphs, which are data structures consisting of nodes and edges, have become incredibly widespread in today's world. They help to model relationships, and are commonly used in all sorts of everyday applications, from social networks to GPS systems. Graph Neural Networks (GNNs) have become a popular tool for performing inference on graph-structured information. Ever since the initial introduction of graph convolution in GCN \cite{kipf2017semisupervised}, which approximated a localized spectral filter for graphs, researchers have been able to improve on these models to extend their performance and decrease their time and memory requirements. GraphSAGE introduced a layer-wise sampling approach to which allowed for more efficient convolution on larger graphs \cite{hamilton2018inductive}. Techniques like transformers have also been applied to GNNs to great benefit \cite{veličković2018graph}. 
\end{comment}

Modern applications such as social networks \cite{hamilton2018inductive}, product recommendation systems, climate modeling \cite{lam2023graphcast}, and traffic prediction \cite{pmlr-v162-lan22a} are generating increasingly large and complex graphs. Standard deep neural networks (DNNs) struggle to learn from graph structures due to their inability to capture the relational dependencies and topological information intrinsic to graphs. To address these challenges, GNNs have been developed to learn hierarchical representations at both node and graph levels, achieving state-of-the-art performance across various domains, including molecular analysis \cite{Wang_2023}. The fundamental architecture of GNNs revolves around message-passing, where node representations are iteratively updated by aggregating feature information from immediate neighbors and their extended k-hop neighborhoods. However, as GNNs get deeper, the receptive field expands exponentially, leading to the ``neighborhood explosion" problem ~\cite{zeng2019graphsaint} which significantly amplifies computational and memory requirements. To address this problem, numerous sampling algorithms have been proposed. For instance, GraphSAGE \cite{hamilton2018inductive} introduces an inductive approach by sampling a fixed-size subset of neighbors during aggregation, thus reducing computational and memory demands. However, designing optimal sampling strategies remains an active area of research \cite{balın2023layerneighbor, zou2019layerdependent}.

% Graph neural networks (GNNs) have found applications in traffic prediction, where intersections and roads are abstracted to a graph structure \cite{pmlr-v162-lan22a}; molecular analysis, where a graph is created from atoms and their bonds to one another \cite{Wang_2023}; global weather prediction, where the graph structure is constructed from a spherical mesh \cite{lam2023graphcast}; and social networks, where graphs have been already been used as underlying data structures for decades \cite{hamilton2018inductive}. Their powerful leverage of relational knowledge has led them to achieve state-of-the-art performance in all of these aforementioned tasks and many more, but they also have a key weakness.

% Unlike traditional Deep Neural Networks (DNNs) in adjacent fields, GNNs suffer from a unique issue known as ``neighborhood explosion". This issue exists because each convolution layer must aggregate the neighbors of every targeted node. Consequently, as these networks get deeper, they grow exponentially in both size and compute requirements. Making matters worse, graph convolutional networks (GCNs) required the Laplacian of the whole graph. GraphSAGE was able to help solve some of these issues by sampling neighbors and aggregating features using a learned embedding \cite{hamilton2018inductive}, but this growth still presents incredible challenges when training a moderately sized model on a large graph. Research on improving sampling methodology to further alleviate these problems is ongoing \cite{balın2023layerneighbor} \cite{zou2019layerdependent}.

Recent research in GNNs has encountered two primary challenges: enhancing model performance and distributing the training workload in a scalable manner without significantly compromising accuracy. Improving GNN performance has proven difficult, as studies suggest that simply increasing the depth or width of GNNs may not be effective \cite{jaiswal2022old} \cite{li2022training} \cite{zhou2021dirichlet}. On the other hand, GNNs seem to have a complex relationship with the traditional scaling laws \cite{kaplan2020scaling}. Large graphs require distributed training, which presents a fundamental trade-off between communication overhead and model performance. Communication-intensive approaches maintain high model accuracy by synchronizing gradients and message-passing operations across partitions but suffer from significant latency and bandwidth constraints. On the other hand, zero-communication strategies \cite{ramezani2022learn} eliminate synchronization overhead, but often lead to performance degradation due to the loss of inter-partition information flow. Recent works \cite{sarkar2024massivegnn, hoang2023batchgnn, kaler2021accelerating} have explored intermediate strategies, demonstrating that reducing communication can mitigate overhead while preserving much of the model's predictive performance.

%Large graphs require distributed training, which is hindered by the communication overhead in GNNs due to the message-passing operations across partitions. Some distributed approaches have attempted to reduce communication \cite{ramezani2022learn, sarkar2024massivegnn}, but they often encounter steep performance loss.

% Recently, GNNs have been grappling with two primary issues: extending the performance of models and distributing the workload of GNN training in a scalable fashion without sacrificing too much performance. Extending the performance of GNNs has been difficult, as many recent works seem to indicate that simply deepening or widening GNNs may not be the solution, as this can lead to over-smoothing and information squashing as well as the traditional DNN issues such as overfitting and vanishing gradients \cite{jaiswal2022old} \cite{li2022training} \cite{zhou2021dirichlet}. GNNs seem to have a complex relationship with the traditional scaling laws \cite{kaplan2020scaling}. Large models and graphs have also often necessitated distribution, which is hindered by the intense communication overhead GNNs require. Some distributed approaches have attempted to reduce communication, but they often encounter steep performance loss when doing so. 

Averaging many independently trained models during training could lead to a more generalized model \cite{izmailov2019averaging}. Building on this concept, Model Soups \cite{wortsman2022model} introduce various methods for mixing fine-tuned models that share a common pre-trained baseline. The approach involves training multiple model replicas as \textit{ingredients} completely independently and in parallel. Due to their shared architecture and initialization, these ingredients occupy a similar loss landscape, allowing them to be \textit{mixed} (combined) post-training to create a \textit{soup} -- a single, superior model that outperforms all its ingredients in performance. This strategy circumvents the overparameterization often associated with traditional model aggregation techniques, such as model ensembles, which can become excessively large and complex, even during inference. Graph Ladling \cite{jaiswal2023graph} extends this concept to GNNs. By modifying Model Soups’ greedy souping approach for GNNs, they demonstrate that training multiple models independently from the same random parameter initialization can outperform traditional GNN ensemble methods in terms of accuracy, computational time, and memory efficiency. 

However, Graph Ladling's Greedy Interpolated Soup (GIS) relies on an exhaustive linear search for the best interpolation ratio between the current soup and the next ingredient at every iteration. Such an exhaustive search does not scale well as more ingredients are added and graphs become larger, because it requires significant memory to evaluate each ratio on the entire validation set and significant time to test all ratios. To combat these issues, we propose a \textit{gradient-descent} driven approach to optimize the mixing process and further develop how we can optimize the collection of \textit{orthogonal knowledge} from the ingredients to the final soup. 

In this work, we address two key research questions and provide an in-depth analysis of GNN model soups, exploring the benefits and limitations of various novel souping strategies across multiple graph benchmark datasets and state-of-the-art architectures. \textbf{RQ1:} \textit{Can gradient descent driven souping algorithms outperform traditional greedy interpolation approaches for GNNs?} \textbf{RQ2:} \textit{How can we mitigate the intense memory and computational demands imposed by large-scale graphs?} %\textbf{RQ3:} \textit{Can our souping approaches benefit from distribution}?\\

Our primary contributions are as follows:
\begin{itemize}
    \item Learned Souping for GNNs (LS), a gradient-descent based souping algorithm adapted from an implementation in LLMs that finds more accurate GNN soups in a fraction of the time compared to state-of-the-art methods.
    \item Partition Learned Souping (PLS), a partition-learned variant of the LS strategy that can significantly reduce memory usage for very large-scale graphs and low-memory devices.
    \item An extensive evaluation of LS and PLS using small-scale and large-scale graph benchmark datasets \textit{Flickr, Reddit, ogbn-arxiv, ogbn-products} with multiple GNN architectures \textit{GCN, GraphSAGE, GAT} to evaluate trade-off in memory usage, training time, and model accuracy under varying graph and model sizes.
    \item We show that LS achieves 1.2\% accuracy improvement and 2.1X speedup on the Reddit dataset with the GAT architecture. PLS achieves 76\% reduction in memory usage and a 24.5X speedup on the ogbn-products dataset with the GraphSAGE architecture without degrading accuracy.
\end{itemize}

This paper is structured as follows: Section \ref{sec:background} covers the background on model soups and their application to GNNs. Section \ref{sec:method} presents our methodology. The details of the datasets, models, and experiments are given in Section \ref{sec:experiment}. Experimental results are detailed in Section \ref{sec:results}, followed by a discussion in Section \ref{sec:discuss}. We conclude in Section \ref{sec:conclusion} with future work explored in Section \ref{sec:future}. Our code is publicly available on Github\footnote{\url{https://github.com/Zubes01/Enhanced-Soups-for-Graph-Neural-Networks}}.

\section{Background and Related Works}
In this section, we provide the background knowledge in Distributed GNNs, Model Aggregation, and Model Soups required to understand the context for our approaches.

\label{sec:background}
% \subsection{Distributed GNNs}
% Distributed training approaches have attempted to help alleviate issues involving large models and large graphs by dividing the workload, but this has also introduced issues of its own. Neighborhood explosion means that any distributed approach which leverages graph partitioning must find a way to limit communication between workers. GraphSAINT proposed a method which sampled training nodes and built complete subgraphs instead of sampling neighbors at each layer, which could be used in a distributed manner \cite{zeng2020graphsaint}. LLCG proposed partitioning the graph and distributing the partitions to a set of workers who ignore any connections between them. Model averaging is performed periodically and a global correction step is taken to reduce the error incurred by ignoring connections between partitions \cite{ramezani2022learn}. Many other distributed approaches have been attempted, but reducing communication and synchronization without significant penalty remains an important issue \cite{lin2022characterizing} \cite{lin2023comprehensive}.

\subsection{Model Aggregation}
Machine learning model aggregation has a history spanning over 30 years, beginning with simple linear combinations of outputs to form the first ensembles \cite{WOLPERT1992241}. This was followed by the introduction of bootstrap aggregating, or \textit{bagging}, which enhanced ensemble performance by training models on different subsets of the data in parallel, allowing each to \textit{specialize} \cite{Breiman1996BaggingP} on. Once trained, these models worked together to create a more robust classifier. Boosting \cite{FREUND1997119} was later proposed that trained models in series, where each model focused on correcting the errors made by the previous ensemble, thereby teaching it to learn from its mistakes. This later led to the well-known Adaptive Boosting algorithm \textit{AdaBoost} \cite{Freund1999ASI}. 

Interestingly, recent literature has extended the concept of ensembling to GNNs in various ways. For example, some work suggests using GNNs to perform ensembling for traditional DNNs \cite{10.1145/3580305.3599414}. Specific ensemble techniques have also been developed for GNNs themselves; for instance, \cite{WANG2022346} demonstrated that a stacking ensemble of Graph Convolutional Networks (GCNs) could significantly improve the performance of computer vision models for Autism Spectrum Disorder diagnosis. In 2021 Open Graph Benchmark's (OGB) Large Scale Challenge \cite{hu2021ogblsc}, numerous works \cite{kosasih2021graph, lsncwb, thakoor2023largescale} leveraged ensemble methods to boost performance on large-scale graph datasets. The winning approach for the node prediction dataset, MAG240M, used a weighted bagging ensemble approach \cite{runimp}, highlighting the potential of model aggregation in advancing GNN research.

\cite{math10081300} proposed serial (GEL) and parallel (P-GEL) ensemble learning methods that utilize boosting techniques as well as a DropNode propagation strategy to train robust GNNs. \cite{10.1145/3489517.3530416} proposed an efficient framework for GNN ensemble creation by isolating nodes that are difficult to classify, performing model pruning, and utilizing a diversity-performance balance in the ensemble construction process. Additionally, \cite{wei2023gnnensemble} trained ensemble models on randomly sampled subgraphs to promote diversity, using a voting mechanism to determine the final prediction.

However, model ensembles come with significant costs. The time and memory requirements during both training and inference can increase substantially, depending on the implementation and the number of models included in the ensemble. Some ensemble architectures further complicate this issue by requiring sequential rather than parallel training of candidate models or demanding extensive synchronization during distributed training, severely limiting scalability.

% However, model ensembles can be a particularly expensive approach to extending performance. Depending on the implementation and number of models trained for the ensemble, they can be many more times expensive to run in terms of both time and memory during both training and inference. Some ensemble architectures also require candidate models to be trained in sequence, rather than parallel, or require significant synchronization during distributed training, posing a huge blow to scalability.

\subsection{Model Soups}
Model soups mitigate the drawbacks of traditional model ensembles while preserving many of their benefits. Once \textit{ingredients} are trained, \textit{soup}-ing algorithms combine the ingredients' parameters into a single model. As a result, model soups do not incur any additional time or memory costs during inference. One of the first souping algorithms, \textit{uniform} souping \cite{wortsman2022model}, averages the parameters of all ingredients. While occasionally effective, this method suffers from not knowing how each ingredient model performs and whether it benefits the soup. Later, a more effective \textit{greedy} souping algorithm (Algorithm \ref{alg:greedy}) was proposed, which sorts ingredients by validation accuracy and iteratively adds them to the soup if they improve accuracy. The soup's weights are averaged across the included ingredients at each step. This technique has since gained popularity for fine-tuning Large Language Models (LLMs) \cite{jang2023personalized} and has been applied in domains such as Adversarial Networks \cite{croce2023seasoning, yang2024adversarial}, LiDAR Segmentation \cite{lidar_segment}, and Image Classification \cite{MARON2022307, liao2023descriptor}.

% Model soups help alleviate the problems of model ensembles while still providing much of the benefits. They work by training many models, termed ``ingredients", from a common initialization without any communication or synchronization. After training is complete, they use a ``souping" algorithm, which combines the model parameters together into a single model. Since all the parameters are combined into a single model, model soups incur no additional time or memory costs at inference time.

% One of the first souping algorithms proposed was uniform souping, proposed in Model Soups \cite{wortsman2022model}. It worked by simply averaging all of the models together. However, Model Soups found that a greedy souping algorithm performed better at the cost of some extra time during the souping stage. Greedy souping worked by first sorting ingredients by their validation accuracy, then iterating through them from highest accuracy to lowest. Upon each iteration, the validation accuracy of the current soup is compared to the validation accuracy of the current soup with the addition of that iteration's ingredient. If the validation accuracy with the ingredient is higher, it is added to the soup. At every point, the soup itself is a model with the same architecture as the candidate ingredients, with its weights set to the average across the weights of the ingredients it contains. The pseudocode for Greedy Souping is shown in Algorithm \ref{alg:greedy} \cite{wortsman2022model}.

\begin{algorithm}[tbh]
    \caption{Greedy Souping}
    \label{alg:greedy}
    \begin{algorithmic}
        \STATE {\bfseries Input:} Candidate ingredients $\mathbf{M} = \{M_{1}, M_{2}, \dots, M_{N}\}$
        \STATE $\mathbf{M}_{\text{sorted}} \leftarrow \texttt{SORT}_{\text{ValAcc}}(\mathbf{M})$
        \STATE soup $\gets \{ \}$
        \FOR{$i = 1$ {\bfseries to} $k$ {\bfseries do}}
            \IF{$\mathsf{\texttt{valAcc}}\left(\mathsf{\texttt{average}}\left( \cup \{ M_{i} \}\right)\right)\geq$ 
            \STATE \ \ \ $\mathsf{\texttt{valAcc}}\left(\mathsf{\texttt{average}}\left(\text{soup}\right)\right)$ }
                \STATE $\text{soup} \gets \text{soup} \cup \{M_{i} \}$
            \ENDIF
        \ENDFOR
        \STATE {\bfseries return} $\mathsf{\texttt{average}}\left(\text{soup}\right)$
    \end{algorithmic}
\end{algorithm}

% Greedy Souping has quickly become a widely-known technique for the fine-tuning of Large Language Models (LLMs) \cite{jang2023personalized}, but has also seen use in various other domains such as Adversarial Networks \cite{croce2023seasoning} \cite{yang2024adversarial}, LiDAR Segmentation \cite{lidar_segment}, and Image Classification \cite{MARON2022307} \cite{liao2023descriptor}.

Greedy Interpolated Souping (GIS) (Algorithm \ref{alg:gis}), is an adaptation of Greedy Souping introduced in Graph Ladling \cite{jaiswal2023graph}. Instead of averaging all ingredients, GIS iterates through a set of interpolation ratios to find the optimal mix between the soup and each new ingredient. Graph Ladling found that GNN-ensemble-level scores could be achieved by training multiple models independently in parallel from the same random initialization and mixing them together using GIS. However, this algorithm relies on an exhaustive search through the interpolation ratios for each ingredient and lacks a mechanism for controlling memory usage. Although traditional minibatching can be utilized with GIS, it further extends the execution time. 

\begin{algorithm}[tbh]
\caption{Greedy Interpolated Souping}
\label{alg:gis}
\begin{algorithmic}
    \STATE {\bfseries Input:} Candidate Ingredients $\mathbf{M} = \{M_{1}, M_{2}, \dots, M_{N}\}$,
    \STATE \ \ \ \ \ \ \ \ \ granularity $g$
    \STATE $\mathbf{M}_{\text{sorted}} \leftarrow \texttt{SORT}_{\text{ValAcc}}(\mathbf{M})$
    \STATE soup $\leftarrow \mathbf{M}_{\text{sorted}}[0]$
    \FOR{$i=1$ {\bfseries to} $K$}
        \FOR{$\alpha$ {\bfseries in} $\texttt{linspace}(0,1,g)$}
            \IF{\texttt{valAcc}(\texttt{interpolate}(soup, $M_i$, $\alpha$)) $\geq$ 
            \STATE \ \ \
            \texttt{valAcc}(soup)}
                \STATE soup $\leftarrow$ \texttt{interpolate}(soup, $M_i$, $\alpha$)
            \ENDIF
        \ENDFOR
    \ENDFOR
\end{algorithmic}
\end{algorithm}

% It is possible to use traditional minibatching approaches in tandem with the GIS algorithm while finding the validation accuracy, but doing this extends the execution time even further.

Despite being a relatively new area, model soups are rapidly evolving. \cite{jang2023personalized} found that LLMs could be fine-tuned for several downstream tasks and souped together on the fly to create a personalized LLM.  Similarly, \cite{chen2023fedsoup} found that model interpolation in Federated Learning could enhance both personalization and generalization. To address budget constraints, \cite{menes2024radin} proposed a souping algorithm using an ensemble approximation method to select the best ingredients. Applying model soups to sparse models, \cite{zimmer2024sparse} achieved improved performance. To achieve better performance in CLIP, \cite{fassold2023frankenstein} partitioned ingredients into components before mixing them, using an algorithm inspired by Algorithm \ref{alg:greedy}. Additionally, \cite{jaiswal2023instant} utilized a souping method to generate Lottery Ticket Hypothesis \cite{frankle2019lottery} quality models, showing that instant pruning could create fine-tuned ingredients more efficiently. These recent works open up potential for meaningful integration with GNN training and souping techniques.

% \cite{jaiswal2023instant} was able to generate Lottery Ticket Hypothesis \cite{frankle2019lottery} quality models using a souping procedure in place of the traditional pruning algorithm. In the same work, it was shown that instant pruning could be used to generate a set of fine-tuned ingredients for souping in a much cheaper manner. Many of these recent works could be combined with GNN training and souping techniques in meaningful ways. 

\section{Methodology}
\label{sec:method}
\begin{figure*}[htbp]
  \centering
  \includegraphics[width=\linewidth]{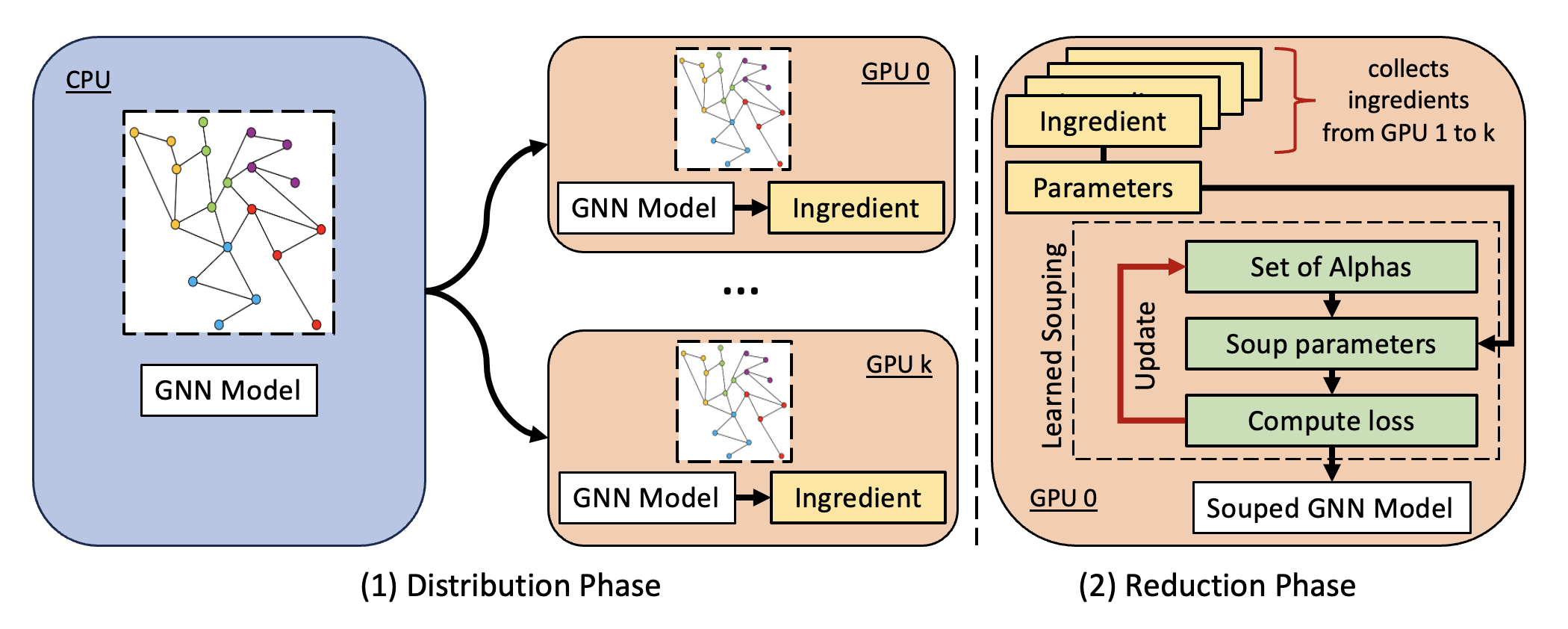}
  \caption{The workflow of our proposed Model Souping in GNNs where Phase (1) \textit{distributes} an initialized model and graph to separate workers, who train independently, becoming \textit{ingredients} and Phase (2) \textit{reduces} the ingredients by \textit{souping} them into a single model.}
  \label{fig:workflow}
\end{figure*}

% In this section, we describe our methodology and workflow for testing our approaches. We briefly explain our approach to training ingredients, implementing Learned Souping, and implementing Partition Learned Souping. We also provide a brief description of the time complexity of our approaches compared to the state-of-the-art. 

Our workflow diagram is shown in Fig. \ref{fig:workflow} and involves training several models concurrently with zero-communication, and then mixing them together. Note that Phase (1) is adapted from the workflow in \cite{jaiswal2023graph}, while our focus is on optimizing performance during Phase (2). In this phase, our Learned Souping and Partition Learned Souping methods \textit{gather} model parameters (``ingredients") onto a single device and mix them together into a single set of model parameters, similar to a \textit{reduce} operation.

\subsection{Distributed Zero-Communication Ingredients Training}
Phase 1 performs a distributed, zero-communication training of the ingredients. The process begins with the user specifying the total number of ingredients $N$, to be trained. A shared model initialization is performed on the CPU and distributed across all the workers. Each worker receives the model as well as the full graph to perform either minibatch or full-batch training. 

To maximize computational resource usage, particularly when the number of ingredients $N>W$ (where $W$ is the number of available workers), a dynamic ingredient training strategy is utilized. Once a worker completes training an ingredient, it immediately begins training the next available ingredient from a shared task queue. This dynamic allocation minimizes idle time across workers and significantly accelerates the ingredient training process. The total training time $T_{total}$ can be approximated as:
\begin{equation}
    T_{\text{total}} \approx \frac{N}{W} T_{\text{single}}
\end{equation}
where $T_{single}$ is the time it takes to train a single ingredient. In an ideal scenario where $ N \leq W$, the minimum training time $T_{min}$ simplifies to 
\begin{equation}
T_{\text{min}} = \max \left\{ T_{\text{single}_i} \right\}_{i=1}^{N}
\end{equation}
This scenario assumes perfect load balancing and uniform ingredient complexity. However, in practice, variability in ingredient complexity may lead to load imbalances, slightly increasing $T_{total}$. Despite this, the process remains embarrassingly parallel, as no inter-worker communication is required during training.
% However, since no communication is required during training, this step is embarrassingly parallel; that is, the minimum training time is just the training time of a single model, provided the training is distributed across enough workers. 

% In order to perform distributed, zero-communication, our approach initialized the model parameters on the CPU before delivering these model parameters to all of the workers. The individual workers then received the entire dataset, and training began. Since no communication is required during training, this step can be exploited by embarrassing parallelism; that is, the minimum training time is just the training time of a single model, provided you distribute the training across enough workers. 

\subsection{Learned Souping}
Learned Souping, originally introduced for LLMs \cite{wortsman2022model}, has not seen widespread use since it requires that all candidate ingredients must be present on the device used for validation. In our work, we adapt this method for GNNs (Algorithm \ref{alg:learnedsouping}), where model sizes are typically much smaller, making it feasible to explore the benefits of Learned Souping in this domain.

The approach introduces a set of interpolation parameters, denoted as $\alpha_{i}^{l}$, for each layer $l$ of the ingredients $M_i$. These parameters control the contribution of each ingredient to the final model. Mathematically, for a model with $L$ layers, the weights of the constructed soup at layer $l$ are defined as:
\begin{equation}
W_{\text{soup}}^{l} = \sum_{i=1}^{N} \alpha_{i}^{l} W_{i}^{l}
\end{equation}

$W_{i}^{l}$ represents the weights of the $i$-th ingredient at layer $l$, $N$ is the total number of ingredients, and $\alpha_{i}^{l}$ are the interpolation parameters. These parameters are initialized and then updated during the training phase.

During training, the soup model $W_{\text{soup}}$ is evaluated on the validation set. The loss, $\mathcal{L}(\text{soup}, G)$, is computed based on the predictions of $W_{\text{soup}}$. The goal is to minimize this validation loss by adjusting the interpolation parameters $\alpha_{i}^{l}$ through backpropagation. At each step, the loss gradient with respect to the interpolation parameters is computed as follows:

\begin{equation}
\frac{\partial \mathcal{L}(\text{soup}, G)}{\partial \alpha_{i}^{l}}
\label{eq:soup}
\end{equation}

The interpolation parameters are updated using Stochastic Gradient Descent (SGD) with a cosine annealing learning rate scheduler. Inspired adversarial networks \cite{huang2023adversarial}, we use Glorot/Xavier initialization \cite{pmlr-v9-glorot10a} for the model parameters and optimize $\alpha_{i}^{l}$ using SGD rather than AdamW commonly used in LLMs. This adaptation ensures compatibility with GNN training dynamics while enhancing performance through fine-tuned learning rate schedules.

% in combination with an SGD optimizer, and a cosine annealing learning rate scheduler. This differs from the original implementation in Model Soups, which used a softmax of 1 over the ingredients (uniform average interpolation) for initialization and the AdamW optimizer without a learning rate scheduler. The process iteratively adjusts the interpolation ratios to find the optimal combination of the ingredients that minimizes the validation loss.

In summary, LS treats the interpolation ratios $\alpha_{i}^{l}$ as learnable parameters, optimizing them through gradient descent to form a single, well-performing model. This differs from simple averaging, as it adaptively learns the best mixture of ingredients to maximize performance on the validation set.

% Learned souping is a souping procedure that was originally introduced in Model Soups \cite{wortsman2022model} but has not seen widespread use, largely because it requires all candidate ingredients to be present upon the device which is used to benchmark performance on the validation set. It works by creating parameters for each ingredient layer, which control the interpolation ratio between all ingredients. The constructed soup is then run against the validation set repeatedly. Each time, loss is backpropagated to the interpolation parameters and adjusted according to an appropriate optimizer. The learned souping algorithm is shown in Algorithm \ref{alg:learnedsouping}.

\begin{algorithm}[tb]
   \caption{Learned Souping}
   \label{alg:learnedsouping}
\begin{algorithmic}
   \STATE {\bfseries Input:} Candidate Ingredients $\mathbf{M} = \{M_{1}, M_{2}, \dots, M_{N}\}$,
   \STATE \ \ \ \ \ \ \ \ \ Epochs $e$,
   \STATE \ \ \ \ \ \ \ \ \ Graph $G$
   \STATE Initialize Alphas using Normal Xavier Initialization
   \FOR{$i = 1$ {\bfseries to} $e$}
       \InlineComment{Multiply each alpha by its corresponding model's layer, then add all of these weights together to build the soup.}
       \STATE $\text{Soup} \leftarrow \texttt{buildSoup}( \mathbf{M}, \text{Alphas} )$ 
       \InlineComment{Generate predictions and calculate loss for the validation set by using a forward pass through the soup}
       \STATE $\text{Loss} \leftarrow \texttt{validationLoss}( \text{Soup}, G ) $
       \InlineComment{Back propagate to update the Alphas}
       \STATE $\text{Alphas} \leftarrow \texttt{updateAlphas}( \text{Loss}, \text{Alphas} )$
   \ENDFOR
   \RETURN $\text{Soup}$
\end{algorithmic}
\end{algorithm}

% This iterative training procedure suffers from a more intensive memory requirement and showcased generally poor performance in the LLM domain when compared with greedy souping \cite{wortsman2022model}. However, Learned Soups have also recently been applied to adversarial neural networks, where they were used to better control the clean-robust tradeoff and achieve higher robustness \cite{huang2023adversarial}. The primary weakness of the approach, requiring all models to be loaded onto the device, has significantly less impact in the domain of Graph Neural Networks, where models tend to be great magnitudes smaller in size. We believe that the memory requirement problem for GNNs is much more a problem of graph size, which can be managed by numerous approaches already developed in the field of GNNs such as subgraph sampling, neighborhood sampling, and minibatching.

% Taking inspiration from its implementation in adversarial networks \cite{huang2023adversarial}, we adjust the Learned Souping by using normal Glorot/Xavier initialization \cite{pmlr-v9-glorot10a} in combination with an SGD optimizer and a cosine annealing learning rate scheduler. This differs from the algorithm used in Model Soups, which used a softmax of 1 over the ingredients (i.e. starting at a uniform average interpolation) for initialization and the AdamW optimizer with no learning rate scheduler.

\subsection{Partition Learned Souping} 
We introduce a novel, GNN-focused approach to learned souping called \textit{Partition Learned Souping (PLS)}, designed to reduce the memory requirements of traditional entire-graph learned souping. PLS utilizes partition sampling to optimize memory usage and computational efficiency while maintaining model performance.

PLS begins by partitioning the graph into a set of $\mathbf{P}$ partitions using a partitioning algorithm such as Metis \cite{metis}, which balances the number of validation nodes across partitions. Formally, the graph $G$ is divided into $\mathbf{P} = \{P_1, P_2, \dots, P_K\}$, where each $P_i$ represents a subgraph. During each epoch, we randomly select $R$ partitions from $\mathbf{P}$ and combine them into a single subgraph, preserving the edges cut during partitioning to retain the graph's structural integrity. This subgraph is then used to compute the activations and loss for the learned souping algorithm.

\begin{figure}[htbp]
  \centering
  \includegraphics[width=0.8\linewidth]{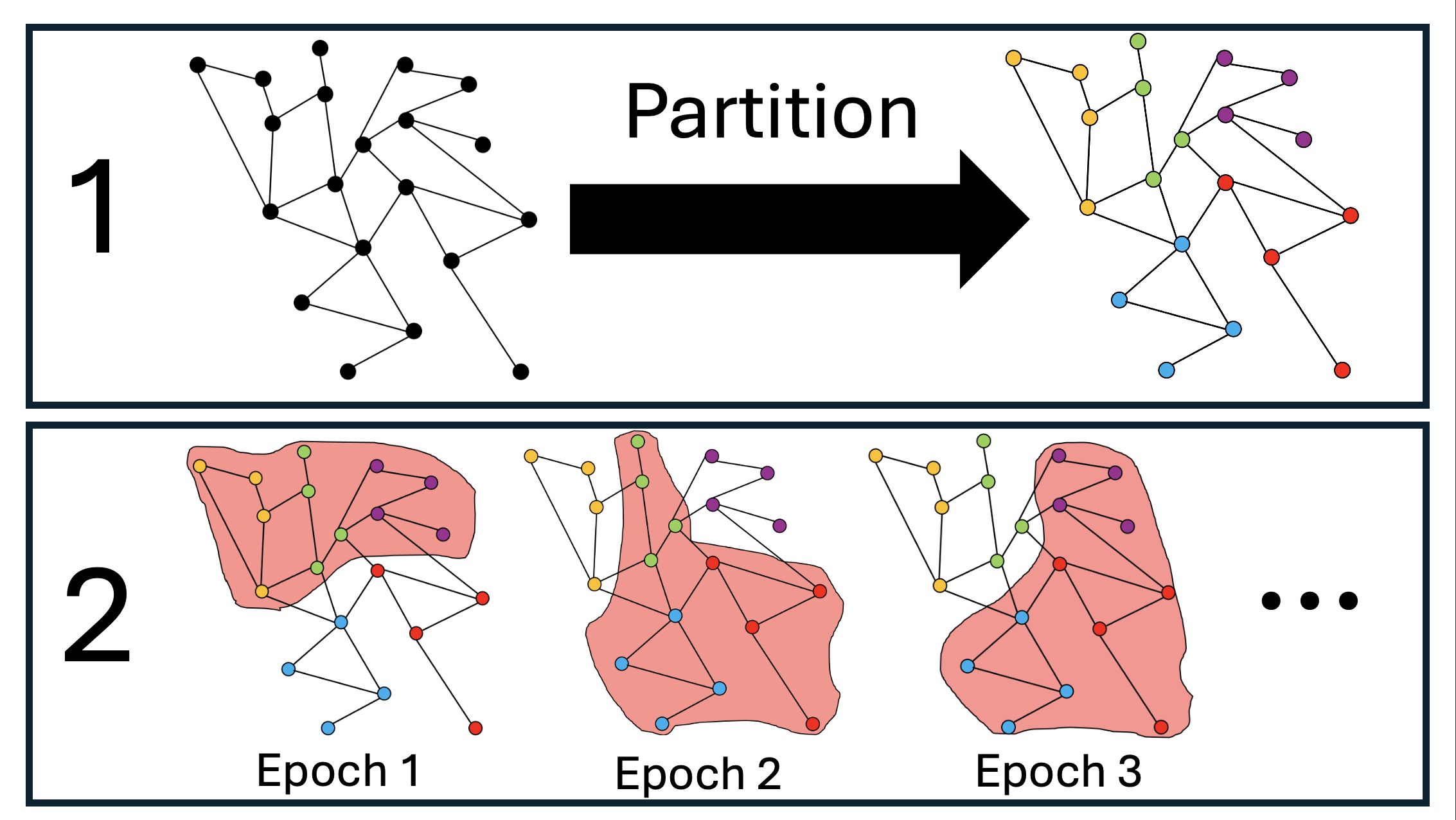}
  \caption{Partition Learned Soup (1) the graph is partitioned as a preprocessing step. (2), during each epoch of Learned Souping, select $R$ partitions to use for training the Learned Soup. }
  \label{fig:pls_visual}
\end{figure}

The core training loop of PLS is summarized in Algorithm \ref{alg:pls} and visualized in Fig. \ref{fig:pls_visual}. At each epoch, we form a subgraph $G_{\text{sub}}$ by selecting $R$ partitions:

\begin{equation}
G_{\text{sub}} = \bigcup_{j=1}^{R} P_{s_j}, \quad s_j \in \{1, 2, \dots, K\}
\end{equation}

$\{s_1, s_2, \dots, s_R\}$ denotes the indices of the selected partitions. The learned soup model is then constructed using interpolation parameters $\alpha_{i}^{l}$, similar to the original learned souping method (Equation \ref{eq:soup}).

After constructing the soup, the model is evaluated on the validation set of $G_{\text{sub}}$. The loss, $\mathcal{L}(\text{soup}, G_{\text{sub}})$, is computed, and the interpolation parameters $\alpha_{i}^{l}$ are updated via backpropagation. The loss gradient for the interpolation parameters is given by:

\begin{equation}
\frac{\partial \mathcal{L}(\text{soup}, G_{\text{sub}})}{\partial \alpha_{i}^{l}}
\end{equation}

This is used to optimize the parameters through an appropriate optimizer. The process is repeated over multiple epochs, gradually refining the soup to maximize its performance. PLS provides greater flexibility in controlling the souping procedure compared to methods like GIS. While GIS relies solely on a single hyperparameter, ``granularity", to adjust the number of interpolation ratios, PLS introduces traditional machine learning controls such as epochs, learning rate, and weight decay, allowing for fine-tuning of the training process. Additionally, the partition selection ratio $R/K$ can be adjusted to control memory usage and training speed, offering a trade-off between resource consumption and model accuracy. In summary, Partition Learned Souping effectively manages memory constraints while retaining high flexibility in the training process, making it a scalable alternative for GNN model souping.

% Our novel, GNN-focused approach to learned souping uses partition sampling to reduce the memory requirement imposed by entire-graph learned souping. We call this approach Partition Learned Souping (PLS). It works by partitioning the graph into $\mathbf{P}$ partitions as a preprocessing step using a partitioning algorithm such as Metis \cite{metis}, which can balance the number of validation nodes in each partition, and selecting $R$ partitions of this graph every epoch. The $R$ partitions are then combined into a single subgraph, adding in any edges cut by the partitioning algorithm, and the validation set of the subgraph is then used to compute activations and loss for the learned souping algorithm. Our approach is visualized in Fig. \ref{fig:pls_visual} and shown in Algorithm \ref{alg:pls}.

\begin{algorithm}[tb]
   \caption{Partition Learned Souping}
   \label{alg:pls}
\begin{algorithmic}
   \STATE {\bfseries Input:} Candidate Ingredients $\mathbf{M} = \{M_{1}, M_{2}, \dots, M_{N}\}$,
   \STATE \ \ \ \ \ \ \ \ \ Epochs $e$,
   \STATE \ \ \ \ \ \ \ \ \ Graph Partitions $\mathbf{P} = \{P_{1}, P_{2}, \dots, P_{K}\}$,
   \STATE \ \ \ \ \ \ \ \ \ Partition Budget $R$
   \STATE Initialize Alphas using Normal Xavier Initialization
   \FOR{$i = 1$ {\bfseries to} $e$}
        \InlineComment{Select R random partitions, which are then joined together into a subgraph, to train on}
       \STATE $\text{Subgraph} \leftarrow \texttt{partitionSelection}(\mathbf{P}, R)$
       \InlineComment{Multiply each alpha by its corresponding model's layer, then add all of these weights together to build the soup.}
       \STATE $\text{Soup} \leftarrow \texttt{buildSoup}( \mathbf{M}, \text{Alphas} )$
       \InlineComment{Generate predictions and calculate loss for the subgraph's validation set by using a forward pass through the soup}
       \STATE $\text{Loss} \leftarrow \texttt{validationLoss}( \text{Soup}, 
       \text{Subgraph} ) $
       \InlineComment{Back propagate to update the Alphas}
       \STATE $\text{Alphas} \leftarrow \texttt{updateAlphas}( \text{Loss}, \text{Alphas} )$
   \ENDFOR
   \RETURN $\text{Soup}$
\end{algorithmic}
\end{algorithm}
\begin{figure*}[!h!t]
    \centering
    \begin{minipage}{\textwidth}
        \centering
        \includegraphics[width=0.4\linewidth]{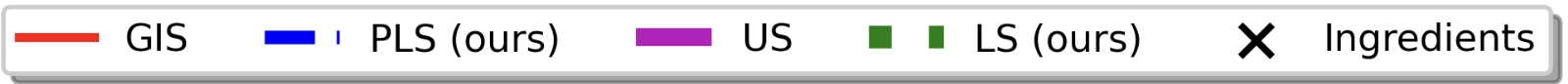}
    \end{minipage}
    \begin{minipage}{0.24\textwidth}
        \includegraphics[width=\linewidth]{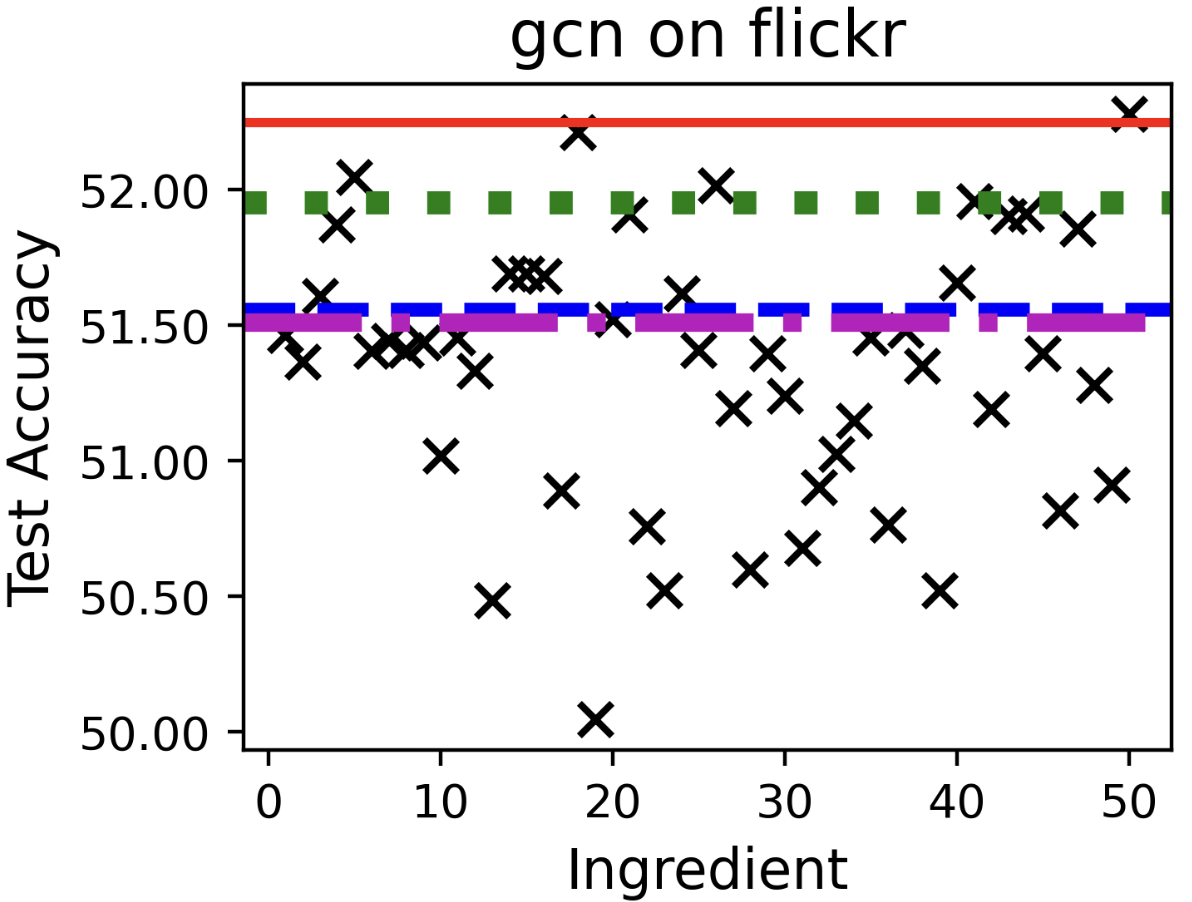}
    \end{minipage}\hfill
    \begin{minipage}{0.24\textwidth}
        \includegraphics[width=\linewidth]{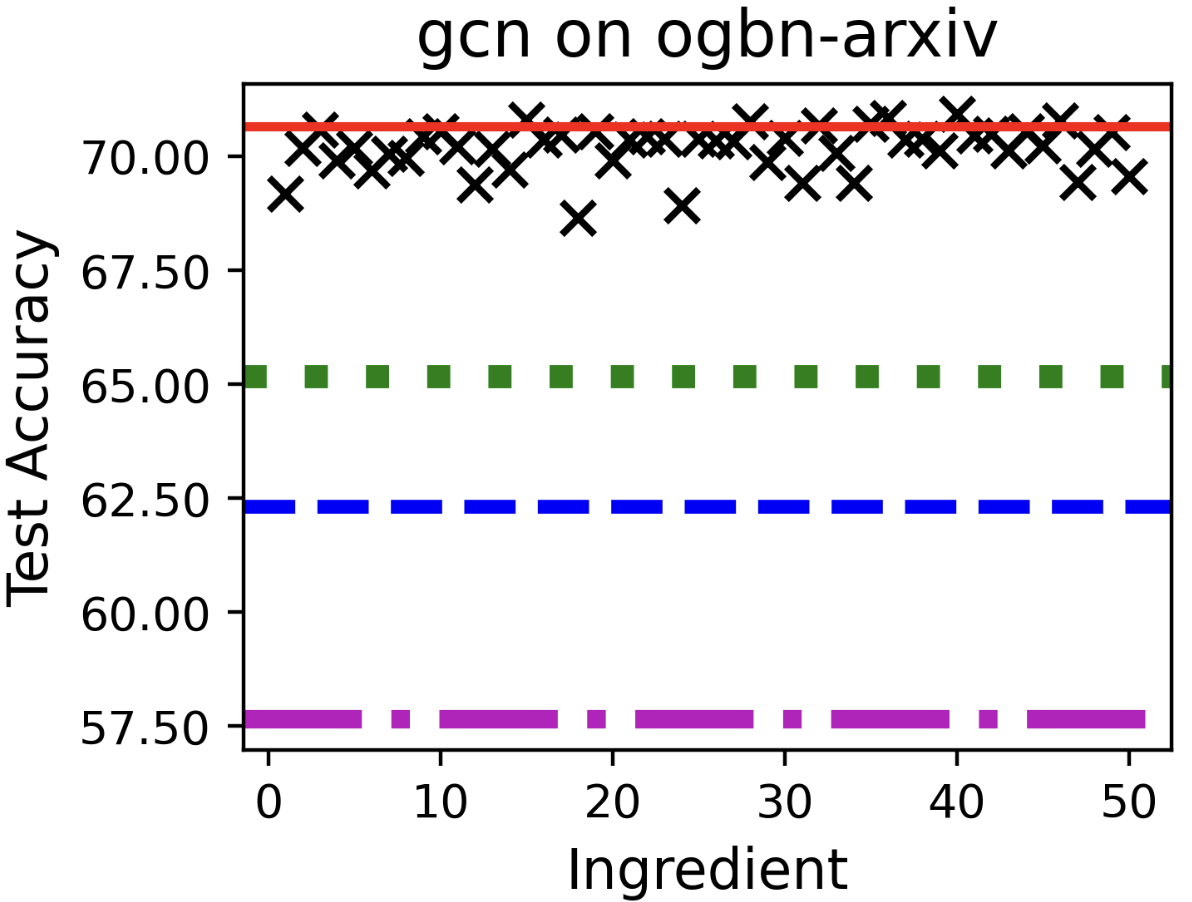}
    \end{minipage}\hfill
    \begin{minipage}{0.24\textwidth}
        \includegraphics[width=\linewidth]{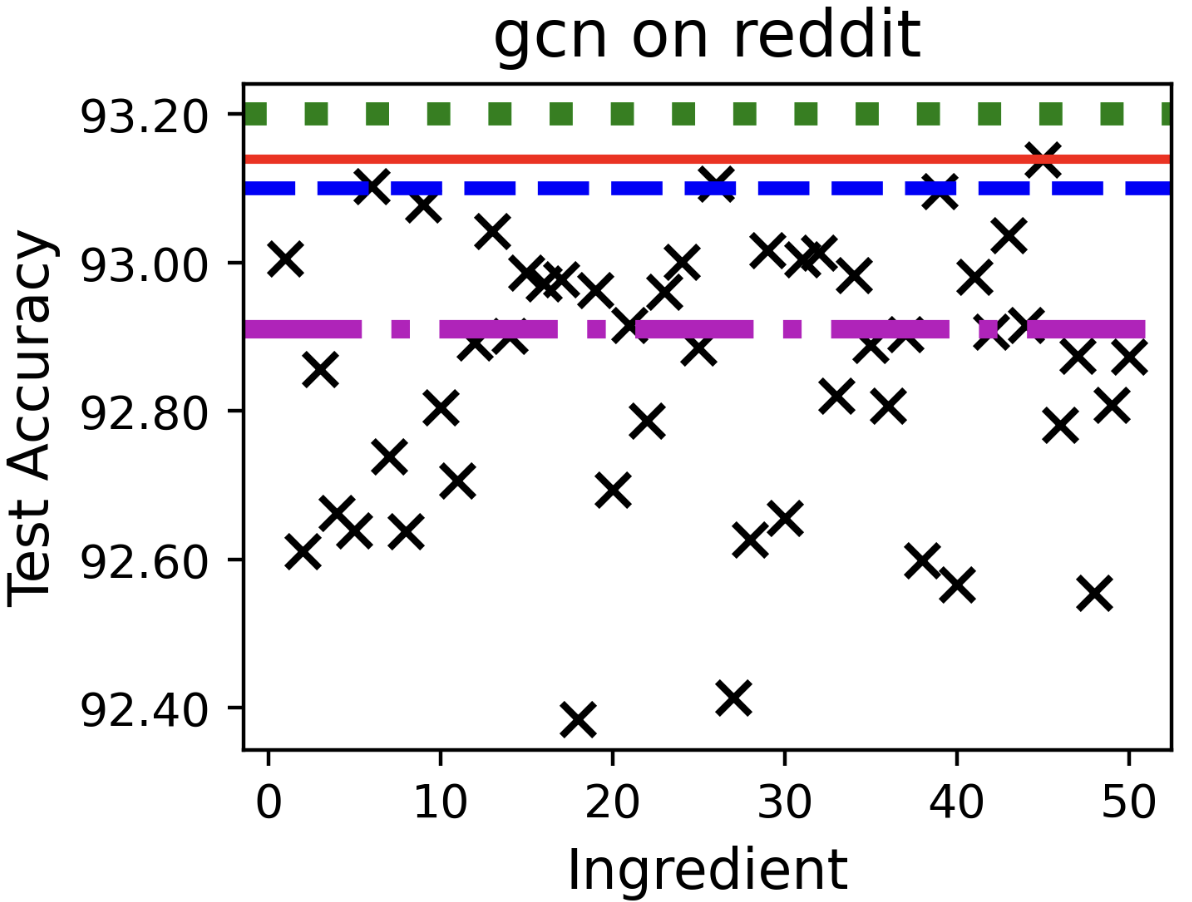}
    \end{minipage}\hfill
    \begin{minipage}{0.24\textwidth}
        \includegraphics[width=\linewidth]{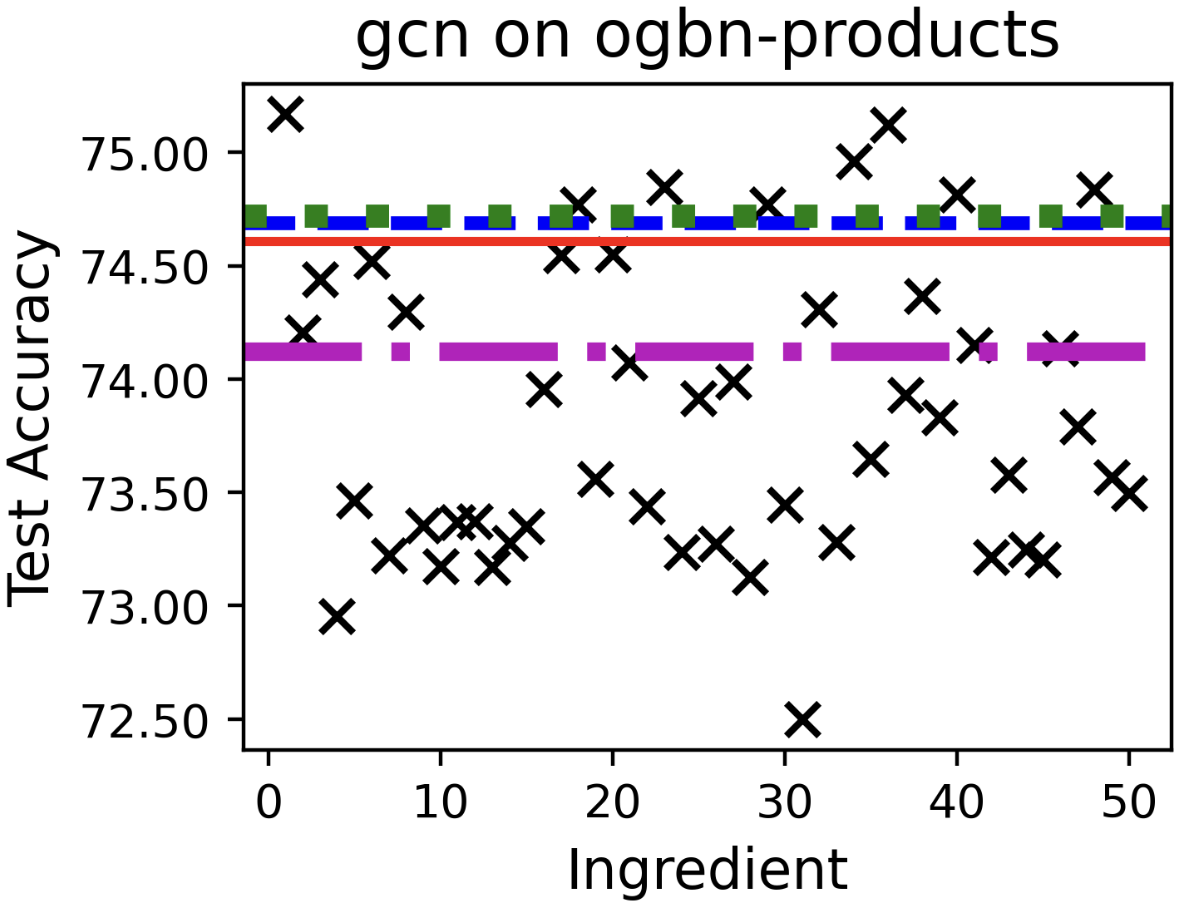}
    \end{minipage}
    \begin{minipage}{0.24\textwidth}
        \includegraphics[width=\linewidth]{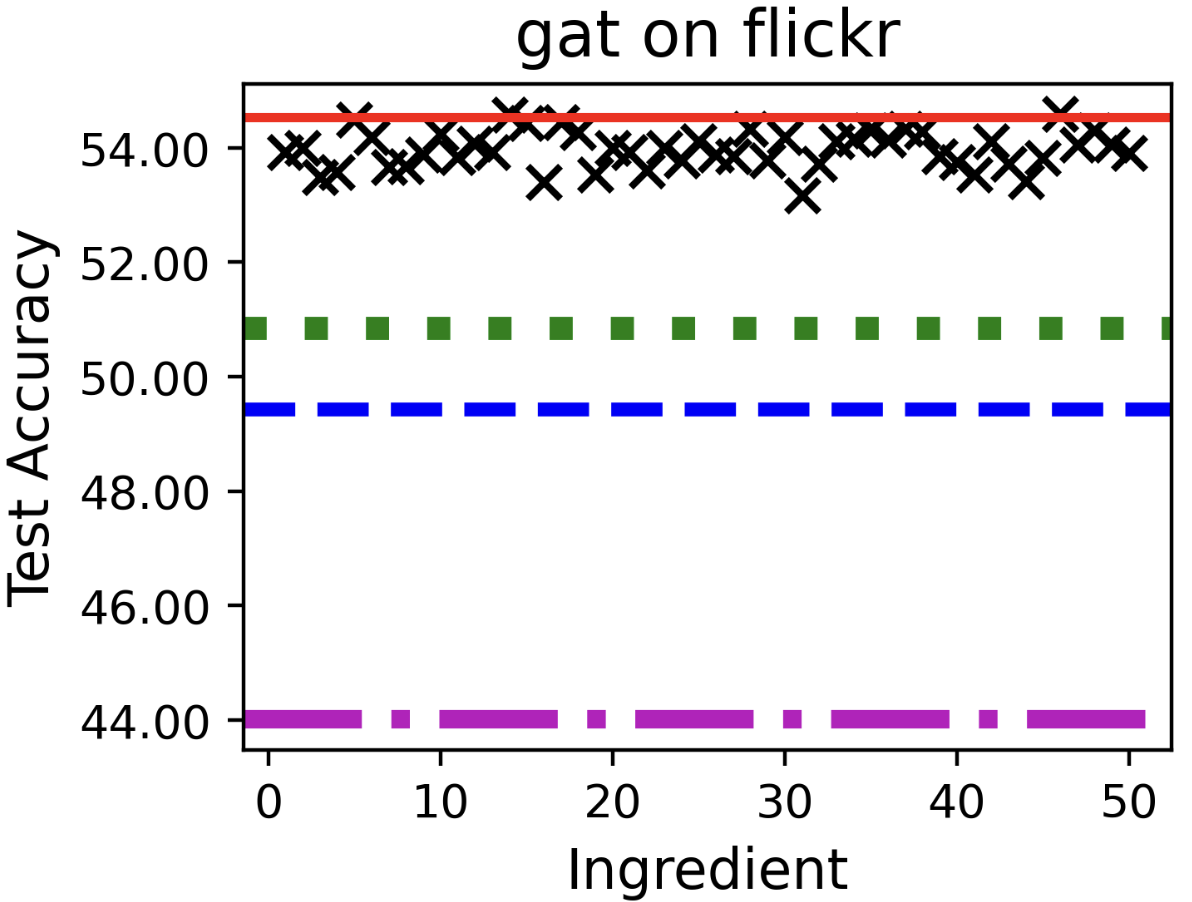}
    \end{minipage}\hfill
    \begin{minipage}{0.24\textwidth}
        \includegraphics[width=\linewidth]{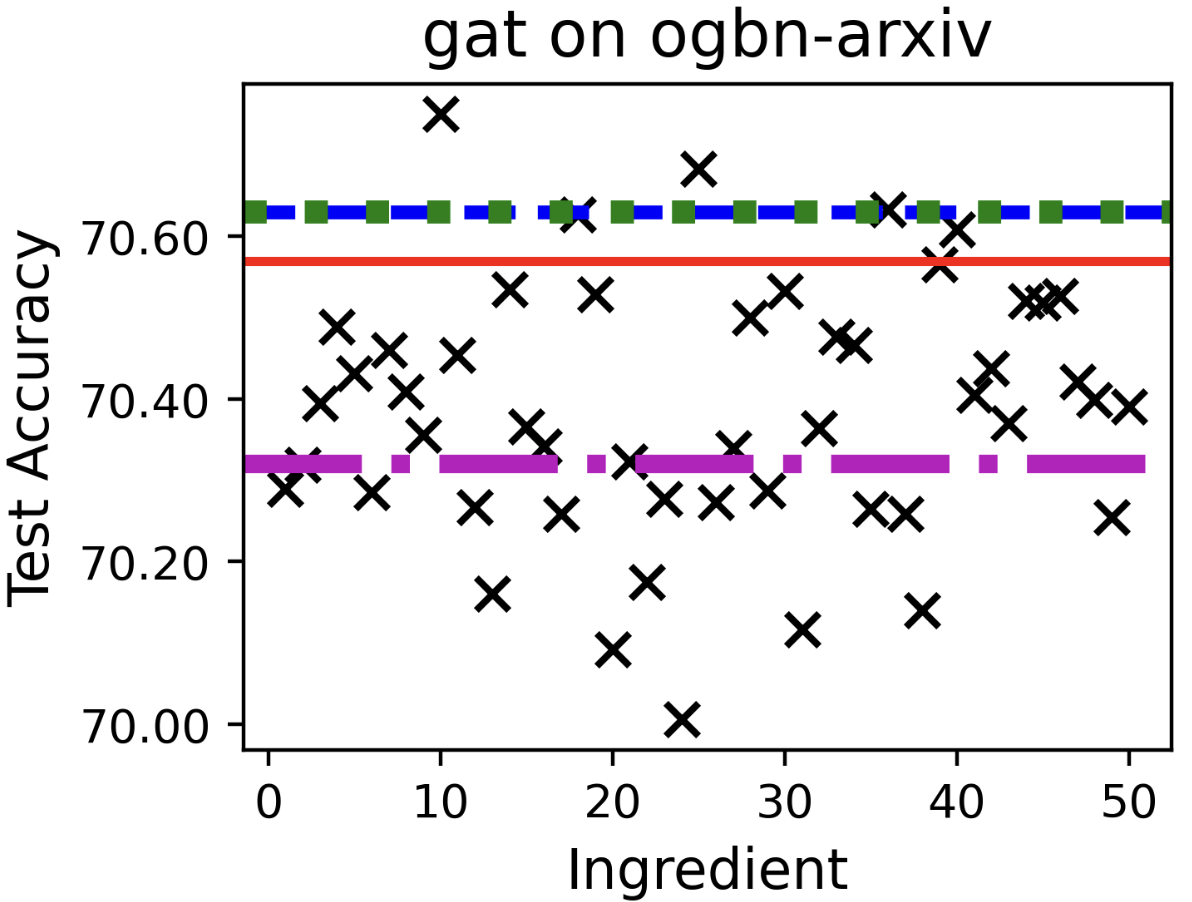}
    \end{minipage}\hfill
    \begin{minipage}{0.24\textwidth}
        \includegraphics[width=\linewidth]{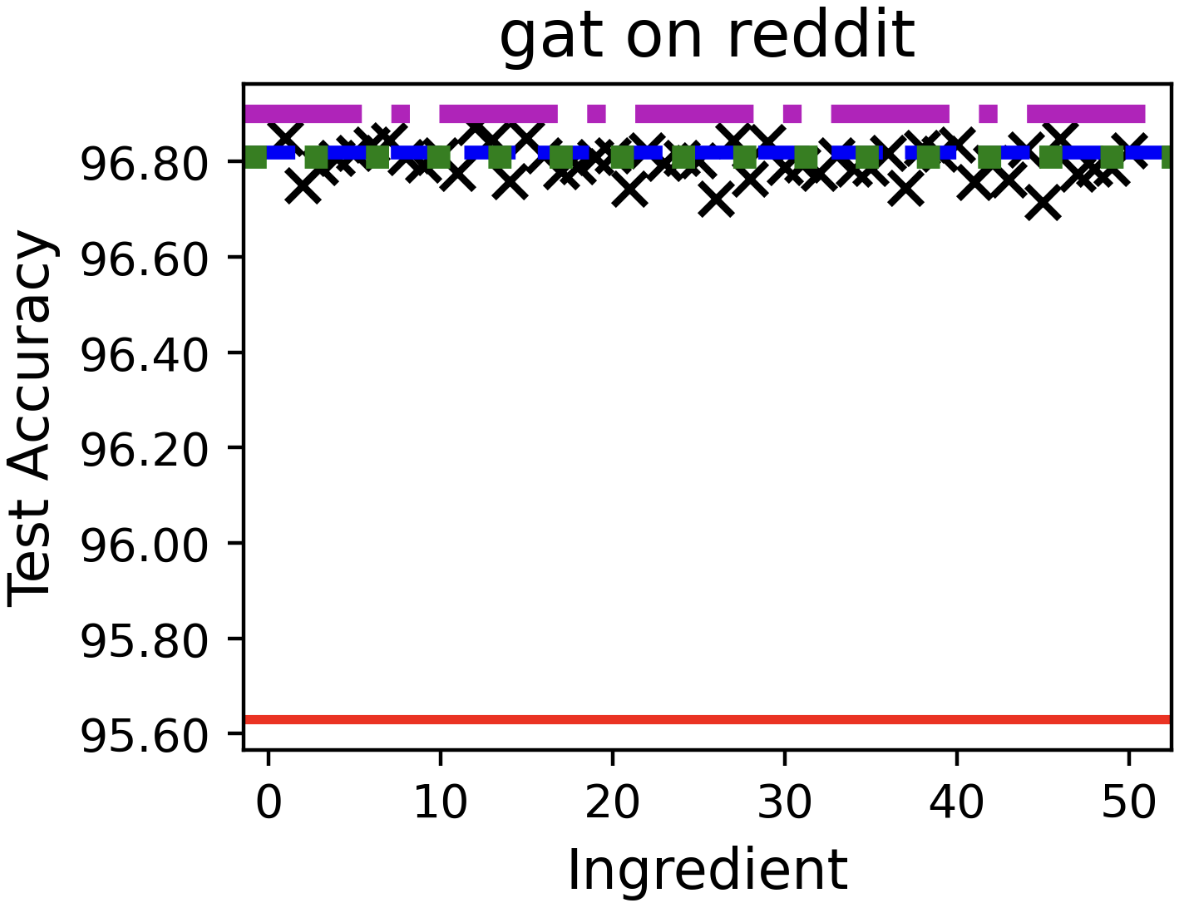}
    \end{minipage}\hfill
    \begin{minipage}{0.24\textwidth}
        \includegraphics[width=\linewidth]{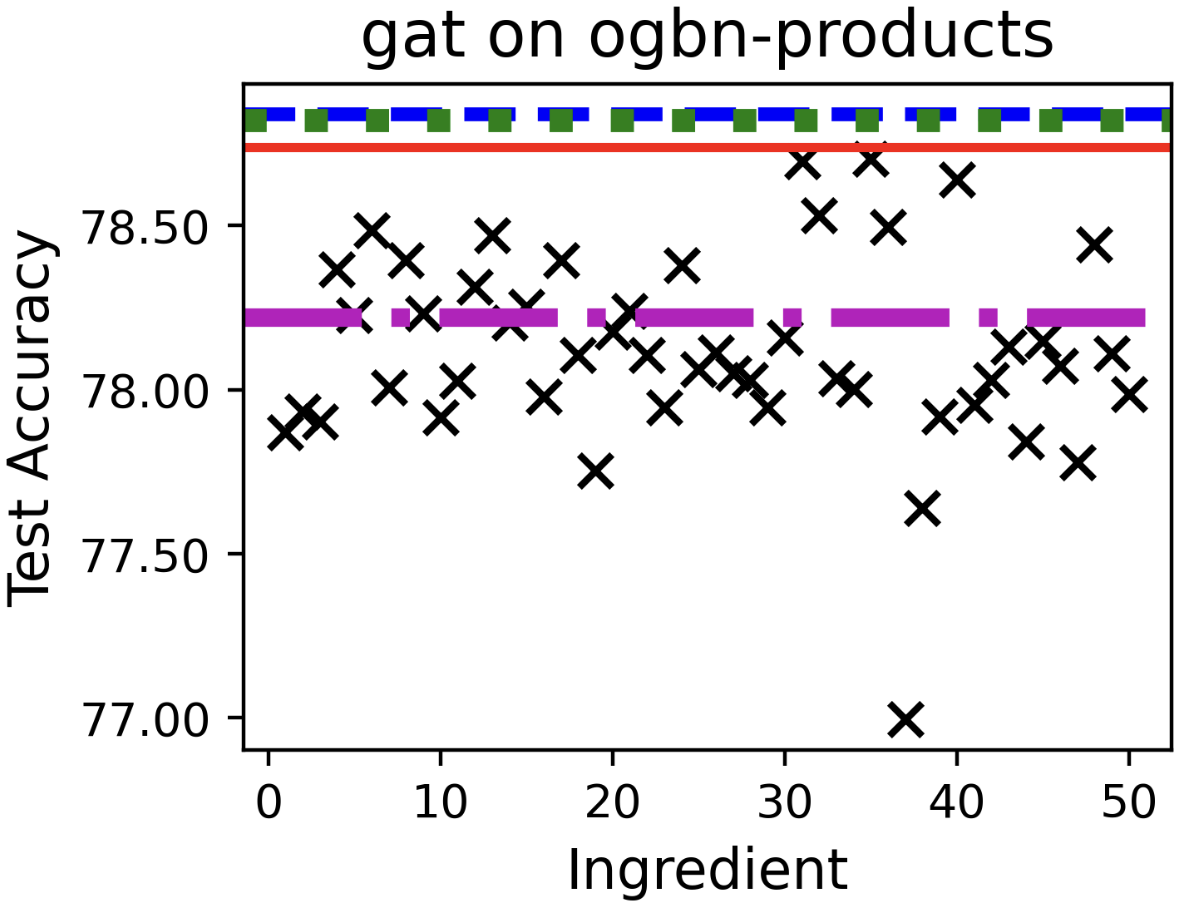}
    \end{minipage}
    \begin{minipage}{0.24\textwidth}
        \includegraphics[width=\linewidth]{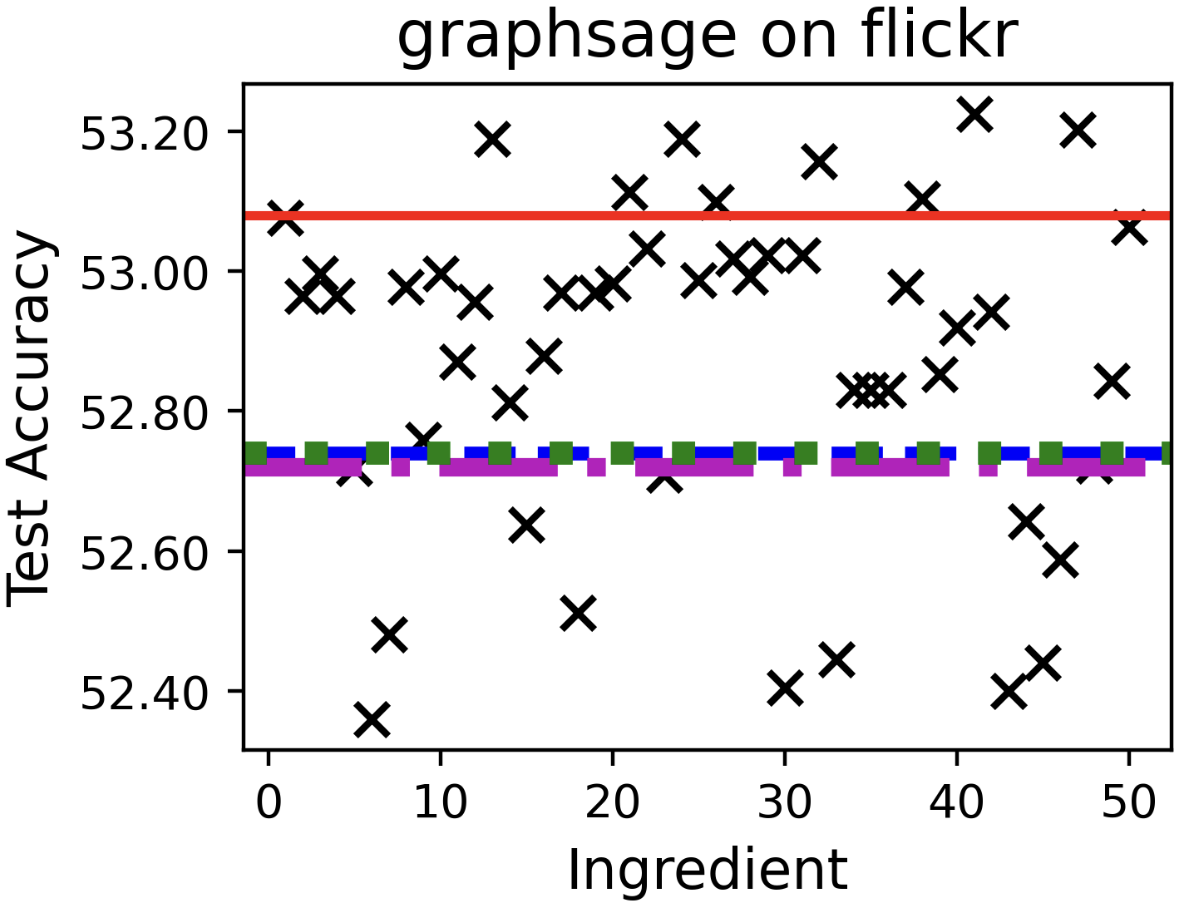}
    \end{minipage}\hfill
    \begin{minipage}{0.24\textwidth}
        \includegraphics[width=\linewidth]{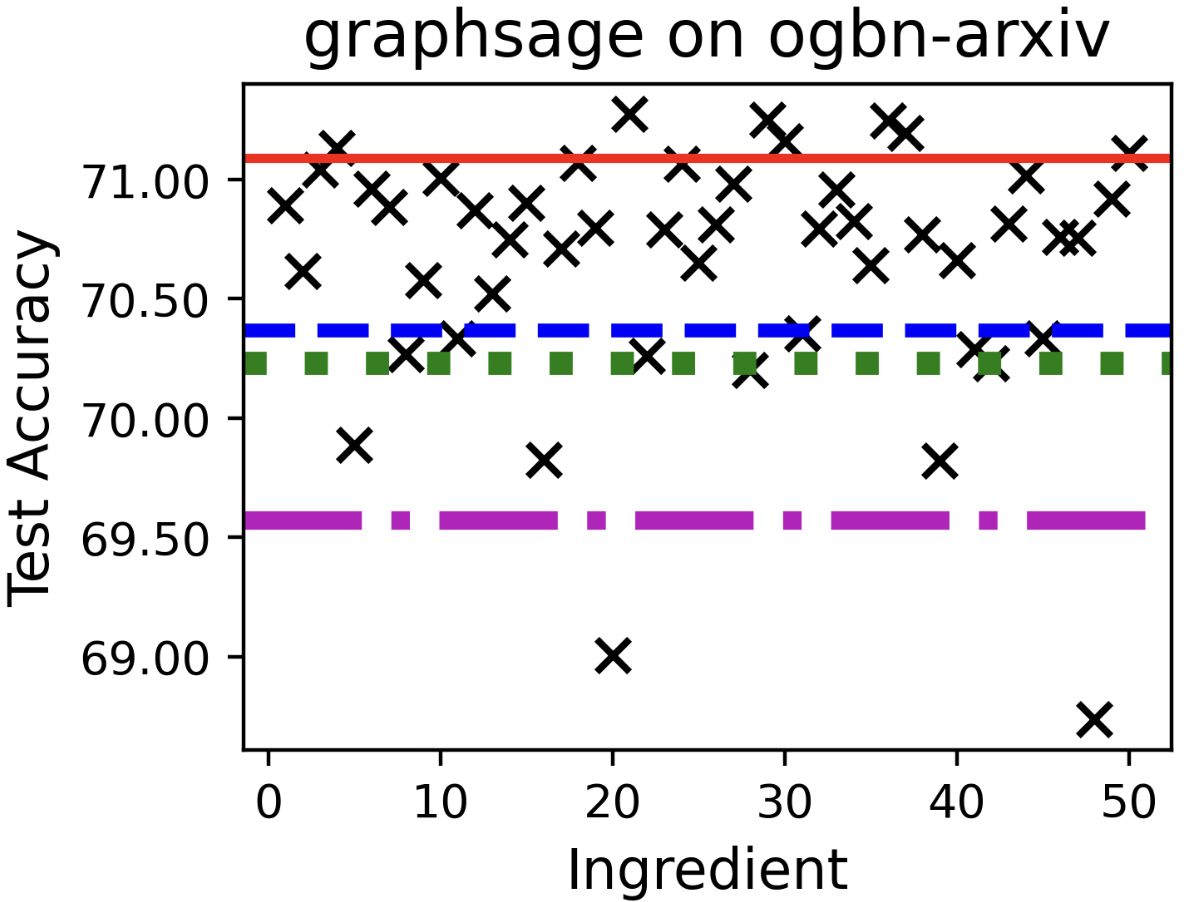}
    \end{minipage}\hfill
    \begin{minipage}{0.24\textwidth}
        \includegraphics[width=\linewidth]{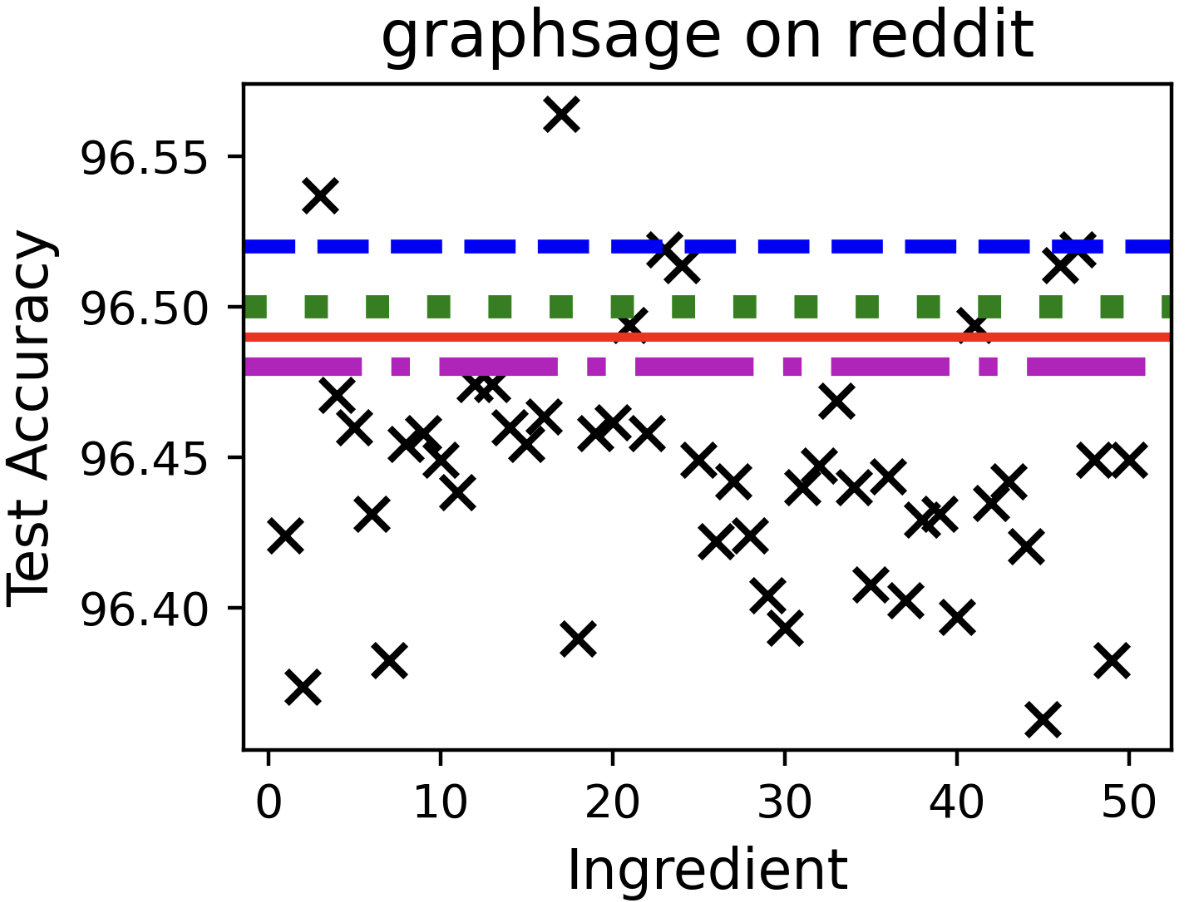}
    \end{minipage}\hfill
    \begin{minipage}{0.24\textwidth}
        \includegraphics[width=\linewidth]{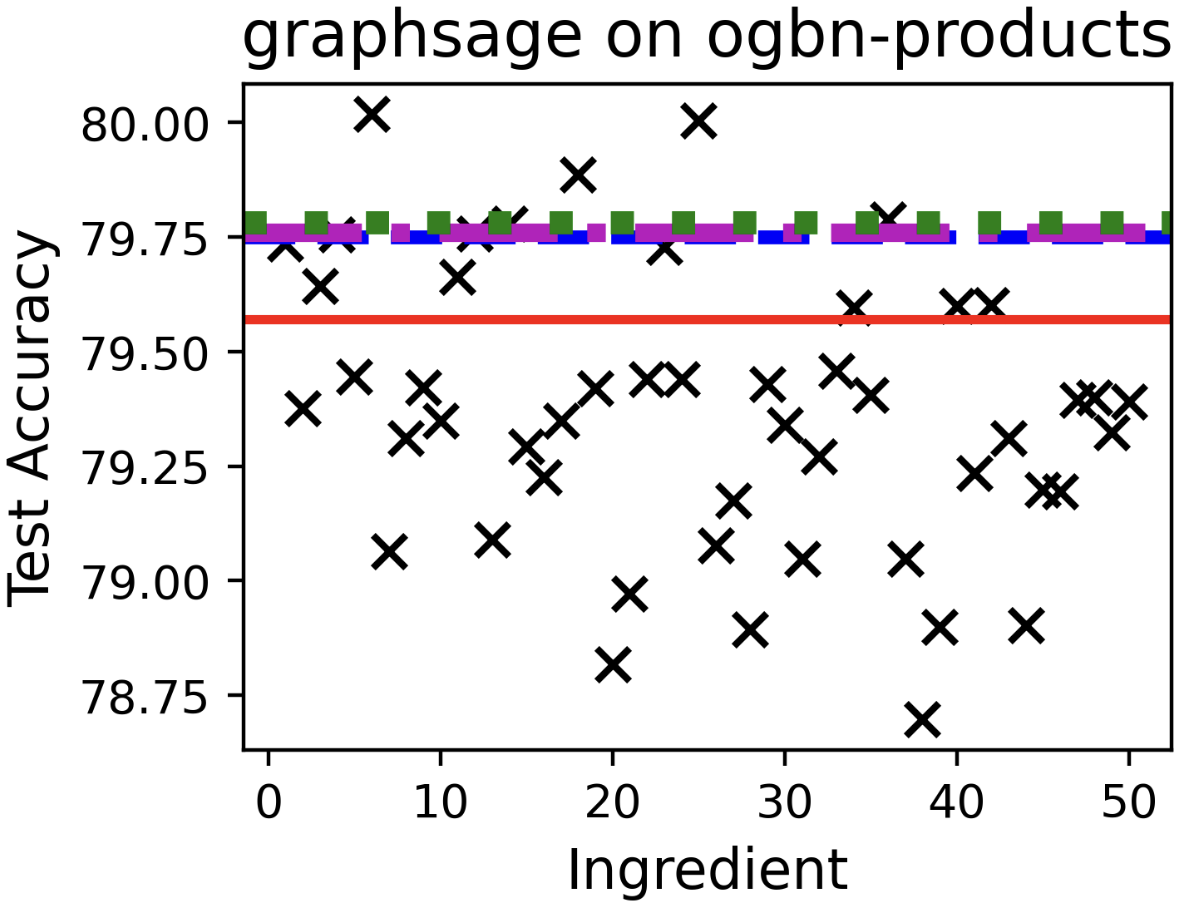}
    \end{minipage}
    \caption{Comparison of different souping strategies across datasets}
    \label{fig:ingredients}
\end{figure*}
% Partition Learned Souping allows us to have much more control over the souping procedure when compared to GIS that uses a single hyperparameter, ``granularity" which controls the number of interpolation ratios which are iterated through. By adjusting granularity, the time taken by GIS can be adjusted at the expense of the final soup performance. On the other hand, Partition Learned Souping has all of the flexibility that we are used to in machine learning: epochs, learning rate, and weight decay can all be adjusted to maximize performance within a desired time constraint. Additionally, the partition selection ratio can be adjusted to decrease memory usage and increase speed at the cost of accuracy.

\subsection{PLS Partition Ratio} 
The \textit{PLS Partition Ratio} defines the fraction of partitions selected to construct a subgraph relative to the total number of partitions $K$ into which the graph was divided. For instance, in the GCN model on the Flickr dataset, the graph was partitioned into 32 parts, and during each epoch of PLS, a new subgraph was constructed using 8 randomly selected partitions, yielding a partition ratio of $R/K = 8/32$.

\subsection{Time Complexity}
The GIS approach iterates over a set of $N$ ingredients $\mathbf{M}$ and a set of chosen interpolation ratios (IR) set by the granularity parameter $g$. For each ingredient ${M_i}$, the approach performs exactly $g$ forward passes on the entire graph to evaluate the validation performance. Its time complexity is $\bigO(NgF_v)$ where $F_v$ is the time taken for each model-IR combination to perform a forward pass on the validation set of the entire graph. While straightforward, this approach is computationally intensive, especially for large-scale graphs or high granularity settings.

Conversely, LS and PLS introduce a new optimization by employing gradient descent to iteratively select the most optimal interpolation ratio of the ingredients, avoiding the exhaustive search inherent in GIS. LS uses an iterative process where forward ($F$) and backward ($B$) propagations are performed across ingredients and interpolation ratios in parallel with gradient descent. This optimization targets efficient souping of ingredients by adjusting the ratios to minimize a predefined loss function. The time complexity for LS is $\bigO(e(F_v+B_v))$ where $e$ is the number of iterations (or epochs). This method converges more efficiently to the optimal interpolation ratios than the brute-force GIS approach. 

PLS, on the other hand, follows a two-fold strategy. It first partitions the graph into $K$ partitions and then randomly selects $R$ partitions during each iteration, adding back their cut edges to form subgraphs. Gradient descent is then performed on these subgraphs, which contain a fraction $\frac{R}{{K}}$ of the original graph's nodes, thereby reducing computational overhead. Since partition selection has a time complexity of $\bigO(R)$, the overall time complexity of PLS is $\bigO(e(R+F_{v'}+B_{v'}))$ where $F_{v'}$ and $B_{v'}$ are the time taken for forward and backward passes on the subgraphs, respectively, for all $v'\subset v$.

\section{Experiments and Analysis}
\label{sec:experiment}
All experiments were conducted on an AMD EPYC 7413 24-core processor and 8$\times$ NVIDIA A100 GPUs using SXM4 with NVLink, serving as the distributed workers. For our implementation, we utilized Deep Graph Library (DGL) \cite{wang2020deep} v1.1.2 and PyTorch v2.1.0 with CUDA 12.6 and NCCL 2.20. We evaluate against two existing souping algorithms for GNNs: Uniform Souping (US), an `uninformed' algorithm, and Greedy Interpolated Souping (GIS), an `informed' algorithm. The baselines \cite{jaiswal2023graph} were originally implemented using Pytorch Geometric (PyG) but to ensure a fair comparison, we re-implemented them in DGL, which offers faster performance and much better memory usage (allowing us to use larger graphs) thanks to kernel fusion.

\subsection{Datasets and Models}
\begin{table}[!b]
\centering
\caption{Dataset Details}
\label{tab:dataset_details}
\resizebox{\columnwidth}{!}{
\begin{tabular}{|c|c|c|c|c|}
    \hline
    \textbf{Dataset} & \textbf{Nodes} & \textbf{Edges} & \textbf{Classes} & \textbf{train/val/test split}\\
    \hline
    Flickr \cite{kipf2017semisupervised} & 89.3K & 0.9M & 7 & 0.5/0.25/0.25 \\
    ogbn-arxiv \cite{hu2021open} & 169.3K & 1.2M & 40 & 0.54/0.18/0.28 \\
    Reddit \cite{hamilton2018inductive} & 233K & 11.6M & 41 & 0.66/0.1/0.24 \\
    ogbn-products \cite{hu2021open} & 2.4M & 61.9M & 47 & 0.1/0.02/0.88 \\
    %ogbn-papers100M \cite{hu2021open} & 111M & 1.6B & 172 & 0.78/0.08/0.14 \\
    \hline
    \end{tabular}
}
\end{table}
We evaluated our proposed souping approaches on four benchmark graph datasets, as listed in Table \ref{tab:dataset_details}. To demonstrate the versatility of our methods, we tested them across three popular GNN architectures: GraphSAGE \cite{hamilton2018inductive}, GAT \cite{veličković2018graph}, and GCN \cite{kipf2017semisupervised}. These models were selected due to their widespread use, relatively small size (enabling quick training of a large number of models), and popularity in GNN model aggregation and ensemble research. 

\subsection{Training Setup and Constraints} Each model was trained on all four datasets with hyperparameters tuned, using cross-validation, across various settings, including both minibatching and full-batching. The constraints for training were: (1) the training time for any particular model should be reasonably short (so that 200 models could be trained in time for the experiment), and (2) the required memory must fit within the available CPU and GPU memory. This setup resulted in a diverse set of model architectures, ranging from as small as 1.4 MB (e.g., GCN on ogbn-products) to as large as 230 MB (e.g., GAT on Flickr). Full details on ingredient training hyperparameters are given in our Github repository..

\begin{table*}[!h]
 \centering
 \caption{Accuracy across datasets for various GNN models, detailing performance enhancements of our proposed LS and PLS against GIS and US. Best scores are in \textbf{bold} and second best scores are \uuline{double underlined}. [Higher is better]}
 \label{tab:souping_accs}
 \begin{tabular}{|c|c|c|c|c|c|c|}
    \hline
    \textbf{Model} & \textbf{Dataset} & \textbf{Ingredients} & \textbf{US} & \textbf{GIS} & \textbf{LS (ours)} & \textbf{PLS (ours)}\\
    \hline
    GCN & Flickr & 51.34 $\pm$ 0.60 & 51.51 $\pm$ 0.04 & \textbf{52.25 $\pm$ 0.15} & \uuline{51.95 $\pm$ 0.09} & 51.56 $\pm$ 0.05 \\
    & ogbn-arxiv & 70.06 $\pm$ 0.60 & 57.65 $\pm$ 0.80 & \textbf{70.64 $\pm$ 0.13} & \uuline{65.17 $\pm$ 1.68} & 62.32 $\pm$ 0.68\\
    & Reddit & 92.85 $\pm$ 0.16 & 92.91 $\pm$ 0.01 & \uuline{93.14 $\pm$ 0.01} & \textbf{93.20 $\pm$ 0.03} & 93.10 $\pm$ 0.03 \\
    & ogbn-products & 73.93 $\pm$ 0.57 & 74.12 $\pm$ 0.08 & 74.61 $\pm$ 0.13 & \textbf{74.72 $\pm$ 0.13} & \uuline{74.69 $\pm$ 0.24} \\
    %& ogbn-papers100M & 60.55 $\pm$ 0.64 & 60.77 $\pm$ 0.05 & 65.84 $\pm$ 0.14 & 64.50 $\pm$ 0.15 & ? $\pm$ ? \\
    \hline
    GAT & Flickr & 54.00 $\pm$ 0.33 & 44.01 $\pm$ 0.23 & \textbf{54.53 $\pm$ 0.21} & \uuline{50.85 $\pm$ 0.10} & 49.43 $\pm$ 0.67\\
    & ogbn-arxiv & 70.37 $\pm$ 0.16 & 70.32 $\pm$ 0.03 & \uuline{70.57 $\pm$ 0.05} & \textbf{70.63 $\pm$ 0.07} & \textbf{70.63 $\pm$ 0.07} \\
    & Reddit & 95.49 $\pm$ 0.06 & \textbf{96.90 $\pm$ 0.01} & 95.63 $\pm$ 0.03 & \uuline{96.81 $\pm$ 0.03} & \uuline{96.82 $\pm$ 0.02} \\
    & ogbn-products & 78.54 $\pm$ 0.27 & 78.22 $\pm$ 0.07 & 78.74 $\pm$ 0.11 & \uuline{78.82 $\pm$ 0.03} & \textbf{78.84 $\pm$ 0.02}\\
    %& ogbn-papers100M & 59.95 $\pm$ 0.62 & 60.87 $\pm$ 0.03 & ? $\pm$ ? & 62.58 $\pm$ 0.37 & ? $\pm$ ? \\
    \hline
    GraphSAGE & Flickr & 52.85 $\pm$ 0.23 & 52.72 $\pm$ 0.03 & \textbf{53.08 $\pm$ 0.03} & \uuline{52.74 $\pm$ 0.04} & \uuline{52.74 $\pm$ 0.03}\\
    & ogbn-arxiv & 70.54 $\pm$ 0.49 & 69.57 $\pm$ 0.25 & \textbf{71.09 $\pm$ 0.16} & 70.23 $\pm$ 0.29 & \uuline{70.37 $\pm$ 0.28} \\
    & Reddit & 96.45 $\pm$ 0.04 & 96.48 $\pm$ 0.01 & \uuline{96.49 $\pm$ 0.02} & \uuline{96.50 $\pm$ 0.01} & \textbf{96.52 $\pm$ 0.02} \\
    & ogbn-products & 79.33 $\pm$ 0.31 & \uuline{79.76 $\pm$ 0.05} & 79.57 $\pm$ 0.096 & \textbf{79.78 $\pm$ 0.04} & \uuline{79.75 $\pm$ 0.05}\\
    %& ogbn-papers100M & 64.64 $\pm$ 0.36 & 66.10 $\pm$ 0.09 & 66.07 $\pm$ 0.05 & 66.47 $\pm$ 0.1 & ? $\pm$ ? \\
    \hline
  \end{tabular}
\end{table*}
\begin{table*}[!h]
 \centering
  \caption{Comparison of time taken (in seconds) by LS and PLS with GIS and US (baseline) across various GNN models and four datasets. Best times are in \textbf{bold} and second best times are \uuline{double underlined}. [Lower is better]}
  \label{tab:souping_times}
  % \resizebox{\columnwidth}{!}{
    \begin{tabular}{|c|c|c|c|c|c|}
    \hline
    \textbf{Model} & \textbf{Dataset} & \textbf{US} & \textbf{GIS} & \textbf{LS (ours)} & \textbf{PLS (ours)}\\
    \hline
    GCN & Flickr & \textbf{8.36 $\pm$ 2.69} & 19.12 $\pm$ 0.03 & \uuline{9.61 $\pm$ 5.22} & 17.24 $\pm$ 5.53\\
    & ogbn-arxiv & \textbf{7.27 $\pm$ 3.38} & 28.63 $\pm$ 0.04 & \uuline{25.65 $\pm$ 5.65} & \uuline{25.05 $\pm$ 5.00} \\
    & Reddit & \textbf{4.76 $\pm$ 0.31} & 326.76 $\pm$ 0.09 & \uuline{65.01 $\pm$ 5.22} & 267.01 $\pm$ 5.20 \\
    & ogbn-products & \textbf{8.95 $\pm$ 3.93} & 437.37 $\pm$ 0.45 & 88.82 $\pm$ 4.79 & \uuline{34.61 $\pm$ 4.99} \\
    %& ogbn-papers100M & 16.57 $\pm$ 9.47 & ? $\pm$ ? & 2247.20 $\pm$ 386.07 & ? $\pm$ ? \\
    \hline
    GAT & Flickr & \uuline{197.48 $\pm$ 8.92} & 738.63 $\pm$ 0.44 & 350.05 $\pm$ 4.37 & \textbf{122.15 $\pm$ 5.89}\\
    & ogbn-arxiv & \textbf{8.57 $\pm$ 2.97} & 114.27 $\pm$ 0.34 & \uuline{37.78 $\pm$ 4.56} & 57.75 $\pm$ 4.45 \\
    & Reddit & \textbf{14.92 $\pm$ 0.53} & 292.73 $\pm$ 1.26 & 137.36 $\pm$ 4.09 & \uuline{38.33 $\pm$ 4.51} \\
    & ogbn-products & \textbf{48.38 $\pm$ 2.01} & 696.47 $\pm$ 2.46 & 533.60 $\pm$ 5.87 & \uuline{70.28 $\pm$ 4.36}\\
    %& ogbn-papers100M & 222.70 $\pm$ 10.10 & ? $\pm$ ? & 851.36 $\pm$ 82.86 & ? $\pm$ ? \\
    \hline
    GraphSAGE & Flickr & \textbf{1.81 $\pm$ 2.93} & 18.25 $\pm$ 0.01 & \uuline{3.60 $\pm$ 5.25} & 5.43 $\pm$ 5.24 \\
    & ogbn-arxiv & \textbf{1.86 $\pm$ 2.88} & 39.73 $\pm$ 0.45 & 30.17 $\pm$ 5.20 & \uuline{19.20 $\pm$ 5.21} \\
    & Reddit & \textbf{5.57 $\pm$ 0.14} & 240.99 $\pm$ 0.02 & 28.92 $\pm$ 3.58 & \uuline{16.83 $\pm$ 5.22} \\
    & ogbn-products & \textbf{6.13 $\pm$ 3.04} & 522.97 $\pm$ 0.57 & 32.90 $\pm$ 4.89 & \uuline{21.37 $\pm$ 5.05} \\
    %& ogbn-papers100M & 7.79 $\pm$ 7.02 & 6710.11 $\pm$ 207.32 & 977.85 $\pm$ 9.07 & ? $\pm$ ? \\
    \hline
  \end{tabular}
  % }
\end{table*}
\subsection{Souping Methods and Hyperparameter Selection} 
We trained 50 ingredients for each combination of model, dataset, and souping method, and reported the results as the average of 4 soups. In total, over 2400 models were trained for these experiments. For LS and PLS, hyperparameters were selected by randomly splitting the validation set for training and validating the soup. These hyperparameters, along with the granularity used for GIS, are given in our Github repository.

\section{Results}
\label{sec:results}
%\todo{update results and discussion with new papers100M information}
In this section, we present our results and highlight our successes in terms of accuracy, time, and memory usage. 

\subsection{Accuracy}
In Table \ref{tab:souping_accs} and Fig. \ref{fig:ingredients}, we show that LS matches or outperforms the state-of-the-art GIS approach in both Reddit and ogbn-products for all three architectures. The most notable accuracy gain is on Reddit with GAT, where PLS beats GIS by 1.2\%. This is attributed to the inherent benefits of learned souping's layer-wise and holistic interpolation strategy. Unlike GIS, which determines interpolation ratios for entire model ingredients, LS optimizes its ratios at the layer level for each ingredient. Additionally, LS considers the weights and ratios of all model's interpolation ratios at once instead of tuning them independently. PLS outperforms LS in GAT on Reddit and ogbn-products, and in GraphSAGE on ogbn-arxiv and Reddit due to its partition selection strategy that mimics minibatching. This introduces regularization and helps in generalizing the soups mixed by PLS.

\textbf{Small graphs}. We observe that our worst performance is on small datasets (Flickr and ogbn-arxiv), particularly in cases where there is significant variation in the performance of individual ingredients. We believe that this is due to the difficulty LS faces in zeroing out the interpolation ratios of poorly performing ingredients while simultaneously maximizing those of the best-performing ones. This is because as the poor-performing ingredients' interpolation ratios near zero, the gradients they produce also shrink considerably, and the softmax function is not able to assign a zero to the interpolation ratio. On ogbn-arxiv with GCN, we observed that GIS often discarded all ingredients except for the one with the highest validation performance. Such a selective strategy is challenging for LS to replicate, as it tends to treat ingredients more equitably during the interpolation process. Fig. \ref{fig:ingredients} shows the performance of the soups compared to the performance of their respective ingredients on the test set.

The uninformed Uniform Souping (US) strategy consistently performed the worst across all datasets and models, except for GAT on Reddit, where it managed to obtain the best score. This is because GAT on Reddit's ingredients were all well-perfoming and uncharacteristically similar (the standard deviation between them was 0.06\%, the smallest of any ingredient set's test performance). This caused the informed methods to overfit, leaving US with the highest accuracy.
\begin{figure*}[!h]
    \centering
    \begin{subfigure}[b]{\textwidth}
        \centering
        \includegraphics[width=0.3\linewidth]{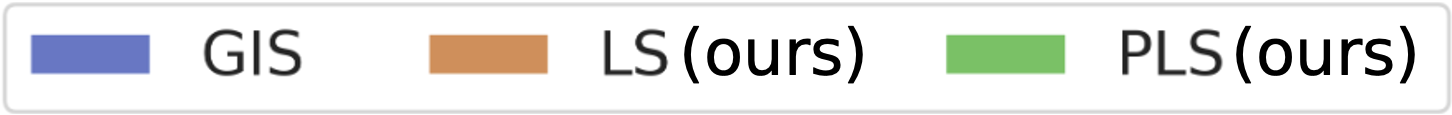}
        \vspace{0cm}
    \end{subfigure}
    \begin{subfigure}[b]{\textwidth}
        \centering
        \includegraphics[width=0.9\linewidth]{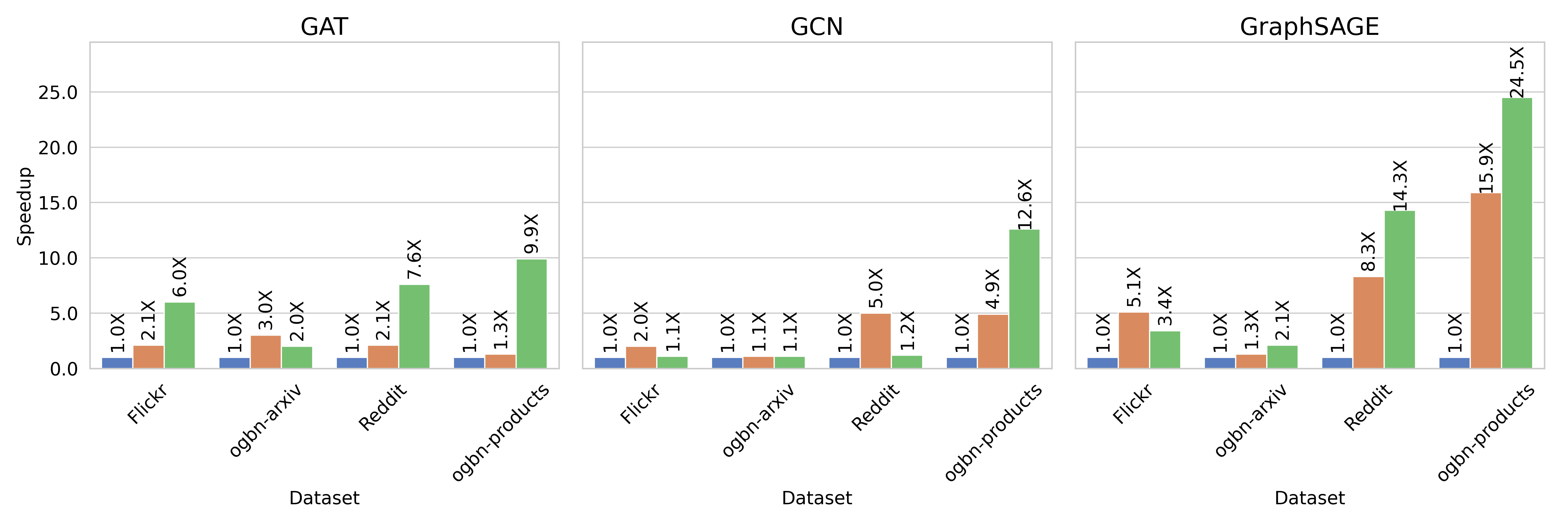}
        \caption{Relative Speedup [Higher is better]} 
        \label{fig:speedupchart} 
    \end{subfigure}
    \begin{subfigure}[b]{\textwidth}
        \centering
        \includegraphics[width=0.9\linewidth]{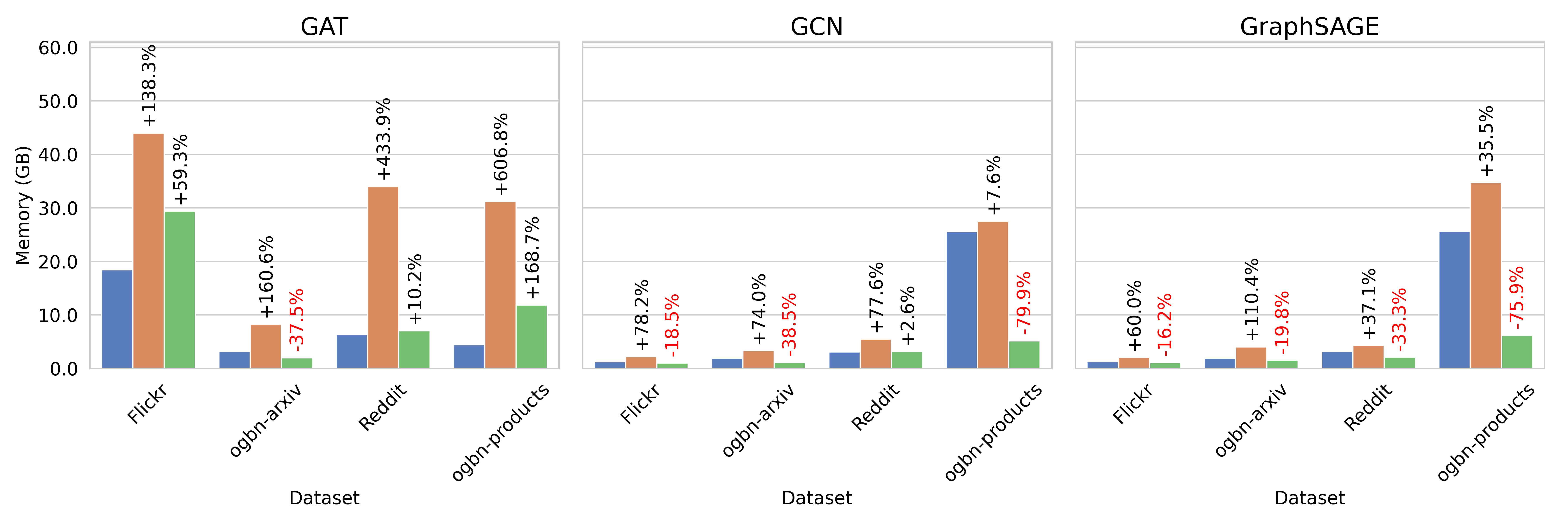}
        \caption{Relative Memory Usage [Lower is better]}  
        \label{fig:memorychart}  
    \end{subfigure}
    \caption{Performance comparison showing (a) Relative Speedup and (b) Relative Memory Usage.}
    \label{fig:performancecomparison} 
\end{figure*}
\subsection{Time}
Table \ref{tab:souping_times} and Figure \ref{fig:speedupchart} show that both LS and PLS achieve faster mixing times than GIS across all datasets and architectures. The highest speedup is observed with the GraphSAGE model on the ogbn-products dataset, where PLS achieves a  24.5$\times$ speedup while utilizing 76\% less memory. This aligns with expectations, as the exhaustive search through interpolation ratios for each model in GIS is inherently more time-consuming compared to the gradient-descent-based learning in LS and PLS. Although PLS achieves faster souping more often, it occasionally requires additional time to reach convergence compared to LS.

Note that the uninformed Uniform Souping (US) strategy nearly always performs best here. This is because, as an uninformed strategy, it does not consider performance on the validation set and only requires taking an average of all of the ingredients' weights. However, as discussed previously, this often results in a significant penalty to accuracy when compared to informed strategies such as GIS, LS, or PLS.

\subsection{Memory}
In Figure \ref{fig:memorychart}, we show the memory impact of LS and PLS. LS demonstrates the highest memory footprint across all 12 dataset-architecture combinations. The relative impact that LS has on memory usage correlates quite well with the number of layers in the model. In contrast, PLS significantly reduces memory usage, achieving the lowest across all datasets for GraphSAGE. The most substantial memory reduction is observed with GCN on ogbn-products, where PLS achieves a 79.86\% reduction in memory usage while also delivering a 12.35 $\times$ speedup without compromising accuracy.

Note that we chose not to include Uniform Souping in this memory usage comparison because Uniform Souping is a completely performance-blind souping algorithm. It does not require any forward passes of the model, meaning that memory usage can be ridiculously low, but as we saw in Table \ref{tab:souping_accs}, this comes at the cost of much worse performance on average.
\section{Discussion}
\label{sec:discuss}
In this section, we explore the findings from our investigations and address each of the research questions raised previously.
\subsection{\textit{Can gradient descent driven souping algorithms outperform traditional greedy interpolation approaches for GNNs?}}
Our results show that gradient descent-driven souping algorithms could outperform greedy interpolation across several model architectures and datasets. As datasets scale, the performance gains from gradient descent-driven souping become more substantial as observed in the Reddit and ogbn-products datasets, where LS outperformed greedy interpolation. Secondly, model size -- specifically the hidden dimension, seems to be an influential factor. This can be seen with the case of GAT on ogbn-arxiv, where a smaller hidden size compared to the other architectures seems to have resulted in a better performance by the learned souping approaches. Despite being faster and more accurate, gradient descent-driven souping presents unique challenges. Optimizing interpolation ratios across layers is non-trivial, with loss landscapes characterized by numerous local minima. This makes both LS and PLS sensitive to hyperparameter settings. Through rigorous cross-validation, we identified hyperparameter settings that achieved speed and performance trade-off. We observed significant performance variability when hyperparameters deviated slightly from these optimal values. Future research could explore techniques like minibatching to stabilize training and mitigate sensitivity. We also observed that relatively large base learning rates often yielded the best results. This likely arises from the ability of larger learning rates to expedite the shrinking of less relevant interpolation ratios toward zero, promoting model sparsity and efficiency. Nevertheless, overfitting remains a reasonable concern, as highlighted in \cite{huang2023adversarial}. Standard techniques to combat overfitting, such as early stopping, may prove valuable in refining learned souping methods.

% Two prevailing factors seem to influence learned souping performance, when compared to GIS performance, above all others: graph dataset size and model size. As graph datasets become larger, the performance of the learned souping approaches seems to increase. This can be seen by observing the consistently improved performance on the Reddit and ogbn-products datasets. These datasets showed consistently better performance for the learned souping approach, even while other datasets used the same model architectures with the same number of layers and hidden dimensions. 

% We were able to successfully perform cross-validation to find hyperparameter settings which we tuned for both speed and performance, and we were able to successfully tune these by hand to adjust the speed-performance trade-off, but we did notice significant swings in performance when hyperparameters changed a bit too much. Techniques such as minibatching could help alleviate some of these issues in future works.

% When finding the optimal base learning rates for our souping procedures, we were surprised to see that relatively large values often worked the best. This is thought to be because large learning rates allow some of these interpolation ratios to more quickly and easily shrink toward zero. Additionally, as touched on in \cite{huang2023adversarial}, over-fitting is a major concern for learned souping. It is possible that existing methods to avoid overfitting, such as early stopping, may be valuable to further research in this area. 

\subsection{\textit{Can we somehow offset the intense memory and time requirements imposed by large graphs?}}
Our PLS approach reduces memory usage by up to 79.9\% and leads to speedups of up to 24.5X, while maintaining comparable accuracy, compared to the greedy interpolated strategy.

PLS relies on the ratio $R/K$, where $K$ is the total number of graph partitions, and $R$ is the number of partitions selected in each epoch. Our experiments show that as model size decreases, the memory reduction achieved by PLS increasingly aligns with this ratio. This is because all models occupy GPU memory, regardless of graph size.

The $R/K$ ratio has interesting implications. Although similar ratios yield comparable memory reductions, they do not always guarantee equivalent performance. When $R$ and $K$ are small, the limited combinations of partitions restrict the variety of subgraphs used in each epoch, leading to degraded performance. In the extreme case of $R = 1$, only 1 partition is selected and no cut edges are used, resulting in information loss.

The number of possible subgraphs is given by the binomial coefficient:
${K \choose R}$. For $R = 1$, this simplifies to $K$, meaning we have only $K$ possible subgraphs, never utilizing the cut edges. Our experiments show that this limited choice can degrade performance by up to 2--3\%.

The objective is to choose values of $K$ and $R$ that are large enough to ensure a unique subgraph in each epoch (i.e., $e \ll {K \choose R}$) while keeping $R$ low to avoid high runtime costs. We found that a practical choice is $R = 8$ and $K = 32$, which results in over ten million possible subgraphs. This large number of combinations ensures a diverse set of subgraphs throughout training, even over a few hundred epochs.

We find that by using a practical $R/K$ ratio, we can reduce memory usage by roughly the same ratio and obtain massive speedups without compromising accuracy.

\section{Conclusion}
\label{sec:conclusion}
In this work, we introduce Learned Souping, a \textit{gradient descent} driven model souping algorithm to the field of Graph Neural Networks. Our investigations reveal that Learned Souping outperforms the state-of-the-art GNN souping algorithm across multiple models and large-scale datasets. We also introduce a novel Partition Learned Souping approach which significantly reduces the memory cost compared to Learned Souping without degrading performance. Both souping methods also significantly reduce the time required to perform souping. We show that Learned Souping reaches an impressive 1.2\% accuracy gain and 2.1X speedup on the Reddit dataset with the GAT architecture. We also show that the Partition Learned Souping approach achieves a memory reduction of 76\% and a 24.5X speedup on the ogbn-products dataset with the GraphSAGE architecture without degrading accuracy. These approaches could be key to extending the performance of Graph Neural Networks without the need to build large and computationally expensive ensembles.

\section{Future Work}
\label{sec:future}
There needs to be a better understanding of the relation between learned souping and traditional gradient descent approaches. It is possible that many existing tools we have developed for Machine Learning are not well suited for the task of mixing model parameters together. Perhaps methods could be used to more easily ``drop-out" poor performing ingredients and help to mitigate the issues faced when using soups on smaller graph datasets. There is also a possibility that the notion of diversity which is known so well in the field of model ensembles could be useful for the preparation of soups.

% \section*{Acknowledgement}
% We would like to thank the Research IT team at Iowa State University for compute resources and support during this research.

\bibliographystyle{IEEEtran}
\bibliography{IEEEabrv,main}

\end{document}